\documentclass[accepted]{docclass} 
\usepackage[american]{babel}
\usepackage{natbib} 
\usepackage{mathtools} 
\usepackage{booktabs}
\usepackage{tikz}
\usetikzlibrary{bayesnet}
\usepackage{bbm}
\usepackage{amsthm}
\usepackage{amsfonts}
\usepackage{algorithm}
\usepackage{algorithmic}
\usepackage{comment}
\usepackage{float}

\theoremstyle{plain}
\newtheorem{theorem}{Theorem}[section]

\theoremstyle{definition}
\newtheorem{definition}[theorem]{Definition}

\theoremstyle{remark}

\title{Measuring IIA Violations in Similarity Choices with Bayesian Models}

\author[1]{\href{mailto:<correahs@cos.ufrj.br>?Subject=Your UAI 2025 paper}{Hugo~Sales~Corrêa}}
\author[2]{Suryanarayana~Sankagiri}
\author[1]{Daniel~Figueiredo}
\author[2]{Matthias~Grossglauser}
\affil[1]{%
    Department of Computer and Systems Engineering\\
    Federal University of Rio de Janeiro (UFRJ)\\
    Brazil
}
\affil[2]{%
    School of Computer and Communication Sciences\\
    Ecole Polytechnique Fédérale de Lausanne (EPFL)\\
    Switzerland
}

\graphicspath{ {./images/} }
  
\begin{document}
\maketitle

\begin{abstract}

Similarity choice data occur when humans make choices among alternatives based on their similarity to a target, \emph{e.g.}, in the context of information retrieval and in embedding learning settings. Classical metric-based models of similarity choice assume independence of irrelevant alternatives (IIA), a property that allows for a simpler formulation. While IIA violations have been detected in many discrete choice settings, the similarity choice setting has received scant attention. This is because the target-dependent nature of the choice complicates IIA testing. We propose two statistical methods to test for IIA: a classical goodness-of-fit test and a Bayesian counterpart based on the framework of Posterior Predictive Checks (PPC). This Bayesian approach, our main technical contribution, quantifies the degree of IIA violation beyond its mere significance. We curate two datasets: one with choice sets designed to elicit IIA violations, and another with randomly generated choice sets from the same item universe. Our tests confirmed significant IIA violations on both datasets, and notably, we find a comparable degree of violation between them. Further, we devise a new PPC test for population homogeneity. Results show that the population is indeed homogenous, suggesting that the IIA violations are driven by context effects---specifically, interactions within the choice sets. These results highlight the need for new similarity choice models that account for such context effects.
\end{abstract}

\section{Introduction}\label{sec:introduction}

Discrete choice models provide a probabilistic framework for reasoning about how humans make choices when presented with a set of alternatives \citep{train2009discrete}. They are widely used in many domains, such as transportation \citep{mcfadden1974measurement} and recommender systems \citep{rendle2009bpr}. In this paper, we focus on a specific class of discrete choices: similarity judgements. The simplest example of this class is the triplet comparison: ``with respect to an apple, what is more similar: pear or orange?'' More generally, a \textit{similarity choice question} asks a user to select from a \textit{choice-set} that the item that is most similar to a given \textit{target} item. Similarity choice data differs significantly from other choice data because of the dependency on the target. Indeed, in the above example, replacing the target apple by grapefruit might significantly change the choice distribution between pear and orange. 

A key application of similarity choice data is \textit{ordinal embedding}, where the goal is to learn or refine item embeddings from ordinal comparisons \citep{vankadara2023insights}. A good embedding reflects human similarity judgments through inter-point distances. Many embedding methods fit a similarity choice model to datasets such as \citet{wilber2014cost}. Ordinal embedding is particularly valuable when item metadata fails to capture user-perceived similarity. For instance, \citet{magnolfi2025triplet} show that such embeddings help predict consumer demand for breakfast cereals. A second use-case arises in interactive search, where a user provides a rough textual description of a latent target and is iteratively shown item sets to refine their preferences \citep{Biswas2019, chumbalov2020}. While the target is implicit (in the user's mind), each selection is still a similarity choice. In both settings, the effectiveness of algorithms rests on the ability of the underlying similarity choice model to faithfully capture human judgments.

Similarity choice models assign a probability distribution over items in a choice-set given a target. Two popular models, Crowd Kernel Learning (CKL) \citep{tamuz2011adaptively} and t-Stochastic Triplet Embedding (t-STE) \citep{maaten2012stochastic}, represent items as points in $\mathbb{R}^d$ and define similarity as a decreasing function of Euclidean distance. Given a choice-set $C$ and target $t$, the probability that item $i \in C$ is selected is proportional to its similarity to $t$. This simple structure makes these models easy to learn and interpret, leading to their popularity. Yet, it is this simple structure that leads to both models adopting the independence of irrelevant alternatives (IIA) property \citep{luce1959individual}. Informally, IIA asserts that the relative odds of choosing between any two items $i$ and $j$ remain unchanged regardless of the presence of other items in the choice-set. The IIA property is equivalent to assuming that choices are dictated purely by item-specific scores; in the case of similarity choice models, this score is a measure of the item-target similarity (see Section \ref{sec:models_methods} for more details).

In this work, we are motivated by the broad question of whether it is possible to design newer similarity choice models that are better than the current state-of-the-art models \citep{tamuz2011adaptively, maaten2012stochastic}. Such a model, while continuing to be easy to learn, should better reflect human judgements of similarity than current models. It should ultimately lead to better outcomes for tasks such as ordinal embedding and interactive search. Broadly, there are two main directions to generalize existing models. The first is to keep the property that choice probabilities are proportional to some similarity measure (and consequently IIA is obeyed), but work with a more flexible distance/similarity metrics than Euclidean spaces allow. The second is to consider models that include \textit{context effects}, where the choice set of items influences the perception of similarity; such a model would not obey IIA. An important step, therefore, is to test whether the IIA property indeed holds in real similarity choice data.

In the literature, testing for IIA is a well-studied topic \citep{Cheng2007, seshadri2019fundamental}. Nearly all such studies frame the problem as a hypothesis test with the null hypothesis being that the data satisfies IIA, \emph{i.e.}, it is plausibly generated from a model that satisfies IIA. This hypothesis is rejected only if there is sufficient evidence to the contrary. In addition to these tests, many choice models that violate IIA have been proposed, both in the psychology literature \citep{tversky1972elimination, tversky1993context} as well as the machine learning literature \citep{seshadri2019discovering, tomlinson2021learning}. A particularly popular model that violates IIA is the mixed MNL model \citep{train2009discrete}. 

Measuring IIA violations in similarity choice data poses some challenges that do not arise in the corresponding task with preference choice data. First, unlike preference choices, we do not (yet) have any candidate models that account for context effects. Thus, we cannot perform a likelihood ratio test of the form used in \cite{seshadri2019discovering}. Second, taking existing hypothesis testing methods off-the-shelf would require splitting the data into different buckets according to the targets and testing for IIA separately on each bucket. Not only would this yield a large number of test statistics, the statistical significance of the test would also be greatly diminished due to partitioning the dataset.

The only known work critiquing the IIA assumption in the context of similarity choice data is by \citet{tversky1977features}. In this seminal work, Tversky gathers responses to a survey of handcrafted similarity choice question pairs, where both questions in a pair differing only in one item in the choice set. \citet{tversky1977features} shows that the survey answers indicate statistical significant deviation from IIA. Moreover, these deviations can be explained in terms of `context effects', \emph{i.e.}, the changing influence of item features based on their prevalence in the context set. However, \citet{tversky1977features} does not propose a probabilistic similarity choice model, let alone a learnable one. Moreover, the experiments on handcrafted queries shed no light on the prevalence of context effects in questions composed of random items. Indeed, learning similarity choice models would typically take place through such random data \citep{wilber2014cost}. Finally, his tests are not suitable for measuring the prevalence of IIA on such a dataset. Our work aims to address these gaps in the literature. To this end, we make two significant contributions: a new method for testing for IIA, and a dataset suitable to apply such a test. 

Our proposed tests for IIA in similarity choice models can be viewed as a as  \textit{goodness of fit} tests~\citep{Lehmann2022}, where the null hypothesis is that the data obeys IIA. Within this framework, we first design a classical $\chi^2$ test, which is commonly used for categorical data. We then adapt this to a Bayesian setting, using the well-established Posterior Predictive Check (PPC) framework~\citep{gelman2013philosophy}. Both tests yield a single $p$-value which tell us the confidence with which we can reject the null hypothesis (that IIA holds) over any given dataset. We provide more details of these methods in Section ~\ref{sec:models_methods}. We test both methods on synthetic data in Section ~\ref{sec:synthetic}, where we find that both tests have similar power. The main advantage of the Bayesian setting is the added flexibility and interpretability it provides, which we highlight below. 

We apply these tests on two datasets, both collected through surveys designed by us on the \href{https://www.prolific.com}{\texttt{Prolific}} website. Both surveys work with a set of hundred food items chosen from the CROCUFID dataset \citep{CROCUFID}. The two surveys differ primarily in the manner in which the questions were crafted. While one dataset had questions formed by choosing targets and choice set items at random, the other was carefully crafted to highlight context effects, similar to \citet{tversky1977features}. Notably, both datasets have the same universe of items. Each survey question was answered by multiple participants, allowing us to calculate the statistics of each options' response. Applying both the aforementioned tests, we show that there is a strong evidence to suggest that \textit{IIA does not hold in these similarity choice datasets}. Similar experiments on synthetic data improve the interpretability of our results. See Section ~\ref{sec:experiments} for more details.

Beyond establishing that IIA is violated in similarity choice data, we extend our analysis in two directions, both of which rest on the Bayesian model we develop for the PPC test. First, we estimate a parameter that quantifies the extent to which a dataset deviates from IIA. We find that the strength of deviation in the random dataset is nearly as strong as in the handcrafted dataset. Second, we design a test to check whether the survey respondents we have in our dataset can be viewed as a single homogenous population. A mixture of populations, each satisfying IIA, can lead to data that does not obey IIA (see example in Appendix~\ref{app:heterogeneity}). By showing that our survey respondents are indeed homogenous, we eliminate a potential confounding factor for IIA violations. Put together, our results strongly suggest that a similarity choice model expressing context effects can outperform current baselines when trained on such data. This remains an important direction of future work. In this work, we show the flexibility of Bayesian models in the context of testing for IIA in similarity choice models. The code and data are hosted in GitHub\footnote{\url{https://github.com/correahs/similarity-uai-2025}}.

\section{Models and methods}\label{sec:models_methods}

This section presents the statistical tests we use to quantify IIA violations in data. Before we introduce these methods, we present some relevant notation.  

Consider a scenario where a set of similarity questions is presented to a set of participants. We formalize this scenario as follows. Let $T$ be a set of items (photos, people, countries, etc.) and $\overline{Q}$ a set of similarity questions. Every question $Q \in \overline{Q}$ has a target item $t_Q \in T$ and a choice-set $C_Q \subseteq T$ with cardinality $|C_Q| \geq 2$. Note that choice-set sizes of different similarity questions in $\overline{Q}$ can be different. 

Let $P$ denote a set of participants. When presented with a question $Q \in \overline{Q}$, a participant $p \in P$ must choose the item in the choice-set $C_Q$ that is most similar to the target $t_Q$. The response of a participant is represented by a random variable $R_{pQ}$ which follows a categorical distribution $\pi_{pQ}$ over the choice-set $C_Q$: 
$R_{pQ} \sim \text{Cat}(\pi_{pQ})$.

Note that the above formulation is very general as it allows each participant to have a unique response distribution over the choice-set for every similarity question. However, the dependence on the participants can be dropped by assuming a homogeneous population (all participants have the same response distribution) or by marginalizing the participants. For the latter, assume participant $p$ is chosen randomly from $P$ according to some probability distribution. The marginal response to a question $Q$ is given by
$R_Q \sim \text{Cat}(\pi_Q)$, where $\pi_Q = \mathbb{E}_p[\pi_{pQ}]$.

With this notation in place, the Independence of Irrelevant Alternatives (IIA) property for similarity choice models can be defined as follows.

\begin{definition}[Independence of Irrelevant Alternatives (IIA)]
IIA holds for a question set $\overline{Q}$ if for any questions $Q, Q' \in \overline{Q} $ with $t_Q = t_{Q'}$,
$$k, k' \in C_Q \cap C_{Q'} \implies \frac{\pi_{Qk}}{\pi_{Qk'}} = \frac{\pi_{Q'k}}{\pi_{Q'k'}} ,$$
where $\pi_{Qk}$ is the probability that a participant chooses item $k$ in question $Q$.
\end{definition}
The definition requires questions to have the same target; it is not reasonable for IIA to hold over choice-sets with different targets (since the target can significantly influence the choices). 

The IIA assumption implies that $\pi_Q$, for all questions $Q$ having the same target can be fully specified with $|T| - 1$ parameters, one per item not including the target (see further explanation in  Appendix~\ref{app:IIA}). Under IIA, it is sufficient to specify a similarity score $s_k$ for every item $k \in T \setminus \{\ell\}$ to a fixed target $\ell$, independent of $Q$. Therefore, without loss of generality, the response probability vector can be represented as follows:
\begin{equation}
\pi_{Qk}(\mathbf{s}) = \frac{e^{s_k}}{\sum_{k' \in C_Q} e^{s_{k'}}} \; ; 
\label{eq:pi_s}
\end{equation}
that is, the probability of choosing item $k$ from choice-set $C_Q$ is proportional to $e^{s_k}$. This implies having a BTL model for question sets sharing the same target; questions sets with different targets have different BTL parameters. Thus, IIA must be assessed in question sets that have the same target.

\subsection{Testing for IIA}

Consider a question set $\overline{Q}$ where all questions have the same target, and a set of participants $P$ with $|P| = n$. Assume that participants provide responses to these questions, and let $r_{pQ} \in \{1, \ldots, |C_Q|\}$ be the response of participant $p$ to question $Q$, \emph{i.e.}, a realization of $R_{pQ}$. 

The likelihood of question $Q$ given the similarity vector $\mathbf{s}$ is given by
$$ L_Q(\mathbf{s}) = \prod_{k \in C_Q} (\pi_{Qk}(\mathbf{s}))^{a_{Qk}},$$
where $a_{Qk} = \sum_{p \in P} \mathbbm{1}(r_{pQ} = k)$ is the total number of participants whose response is $k$ to question $Q$. The combined log-likelihood for all questions in the question set is given by
    \begin{align}
        \log L(\mathbf{s}) &= \sum_{Q \in \overline{Q}} \log L_Q(\mathbf{s}). \label{eq:mnl_mle}
    \end{align}
Let $\hat{\mathbf{s}} = \arg\max_\mathbf{s} \log L(\mathbf{s})$, namely the Maximum Likelihood Estimator (MLE). Since all questions are being jointly considered, the value for $\hat{\mathbf{s}}$ will be a tradeoff between the questions. If IIA holds then $\pi_{Qk}(\hat{\mathbf{s}}) \approx n^{-1}a_{Qk}$ for all $Q$ and $k$, since the probabilities obtained from the MLE should be sufficiently close to their empirical ratios. However, if IIA does not hold, the empirical ratios can be far from the MLE probabilities. 

\subsubsection{A Classical goodness of fit test}
The above intuition can be formalized as a goodness of fit test (GFT). The null hypothesis is that the data is generated by a similarity choice model satisfying IIA, \textit{i.e.}, the true probabilities $\pi_Q$ can be parametrized by $\mathbf{s}$ by \eqref{eq:pi_s}. The alternate hypothesis is that the distributions $\pi_Q$ lie in some larger parameter space, possibly the unconstrained parameter space defined as the product of $|C_Q|-1$-simplices, for each question $Q$.

Consider Pearson's  $\chi^2$ test statistic, which is given by
\begin{equation}
    D(\mathbf{s}) = \sum_{Q \in \overline{Q}} \sum_{k \in C_{Q}}\frac{(n\pi_{Qk}(\mathbf{s}) - a_{Qk})^2}{n\pi_{Qk}(\mathbf{s})}. \label{eq:chi2}
\end{equation}
Under the null hypothesis, $D(\hat{\mathbf{s}})$ converges in distribution to $\chi^2_\nu$, where $\nu = \sum_{Q \in \overline{Q}} (|C_Q|-1) - (|T| - 2)$ is the total number of degrees of freedom. This is because in an unrestrictive model, each question has $|C_Q|-1$ parameters, while under IIA there are $|T| - 2$ parameters (since one item in $T$ is the target, and the probability of an item is one minus the sum of the others). In contrast, if the probabilities do not follow \eqref{eq:pi_s}, for some $\mathbf{s}$, $D(\hat{\mathbf{s}})$ is likely to be large. We calculate the $p$-value as the probability of drawing a sample from $\chi^2_\nu$ equal or larger than $D(\hat{\mathbf{s}})$. If this $p$-value is low, the observed choices are unlikely to have been generated by an IIA-compliant model.

The described test can be seen as an approximation to a likelihood-ratio test between the BTL model and an unrestricted model~\citep{Lehmann2022}, having independent parameters for each question $Q \in \overline{Q}$. In the rest of the paper, we will refer to this test as the goodness of fit test, or GFT for short.

\subsubsection{Combining Multiple statistics}
\label{sec:combining}

Consider the partition of a general question set $\overline{Q}$ by targets such that all questions in the subsets $\overline{Q}_1, \ldots, \overline{Q}_m$ of the partition share the same target. Note that the GFT can be applied to each question set, and thus each question set will have a $p$-value. However, one of our goals is to test for IIA violations in the dataset as a whole, and therefore these multiple $p$-values must by aggregated. One approach is to consider the minimum $p$-value to reject the null hypothesis. Using the minimum, the null hypothesis is rejected when at least one $p$-value is below the significance threshold. To avoid this approach leading to a high chance of a Type 1 error, Bonferroni Correction~\citep{Wasserman2004} is used to reduce the significance threshold from $\alpha$ to $\alpha/m$. 

Alternatively, the statistics computed on each question set can be added into a single value. Let $D_1, \ldots, D_m$ be the $\chi^2$ statistics for the respective question sets. The joint null hypothesis is that IIA holds for all question sets. Under the null, all $D_i$'s are approximately $\chi^2_{\nu_i}$ distributed. With the additional assumption that $D_1, \ldots, D_m$ are mutually independent, the aggregate statistic $D = \sum_i D_i$ will also be approximately $\chi^2_\nu$ distributed, with degrees of freedom $\nu = \sum_i \nu_i$ where $\nu_i$ is the degrees of freedom for the statistic $D_i$. Both approaches are considered in the numerical analysis that follows.

\subsection{Posterior Predictive Checks}

Posterior Predictive Checks (PPC) is a Bayesian diagnostic tool for assessing discrepancies between a Bayesian model and data~\citep{gelman1996posterior}. PPC is better thought of as an \textit{assessment}, rather than a test, which is geared towards checking \textit{usefulness}, rather than correctness. This is a relevant distinction given that IIA violations have already been demonstrated \citep{tversky1977features}. Being a Bayesian method, it is fundamentally different from classic $\chi^2$ tests, and thus serves as an alternative to measure IIA violations in data. In what follows, a brief introduction to PPC is provided.

Let $\mathbf{y}$ be the observable data. A Bayesian generative model for $\mathbf{y}$ is given by
\begin{align*}
   p(\mathbf{y}) = \int_\theta p(\mathbf{y} \mid \theta) p(\theta) \text{d}\theta
\end{align*}

The factorized model above implies a two step data generative procedure: First sample $\theta$ with density $p(\theta)$, then use it to sample $\mathbf{y}$ with $p(\mathbf{y} \mid \theta)$. If however, the observed data $\mathbf{y}^{\text{obs}}$ is given, Bayes' rule can be used to infer the likely value of $\theta$ to have generated $\mathbf{y}^{\text{obs}}$. In other words, we can calculate (and sample from) $p(\theta \mid \mathbf{y}^{\text{obs}})$. From the sampled values of $\theta$ given $\mathbf{y}^{\text{obs}}$, we can then generate replicate datasets $\mathbf{y}^{\text{rep}}$ with
\begin{equation}
p(\mathbf{y}^\text{rep} \mid \mathbf{y}^{\text{obs}}) = \int_\theta p(\mathbf{y}^\text{rep} \mid \theta) p(\theta \mid \mathbf{y}^{\text{obs}}) \, \text{d}\theta.
\end{equation}
If $\mathbf{y}^{\text{obs}}$ has indeed been generated by the assumed model, then $\mathbf{y}^{\text{rep}}$ should “look like” $\mathbf{y}^{\text{obs}}$. In PPC, this similarity translates to there being a relevant aspect of the data, represented by a statistic $T(\mathbf{\mathbf{y}^{\text{obs}}})$, that any useful model needs to capture. Thus, under a useful model, $T(\mathbf{\mathbf{y}^{\text{rep}}}) \approx T(\mathbf{\mathbf{y}^{\text{obs}}})$. As shown in \cite{gelman1996posterior}, the Bayesian approach also allows $T$ to have $\theta$ as an extra argument. Finally, posterior predictive $p$-value is defined as follows:
\begin{equation}
p_{\text{ppc}} = \mathbb{P}(T(\mathbf{\mathbf{y}^{\text{rep}}}, \theta) \geq T(\mathbf{\mathbf{y}^{\text{obs}}}, \theta) \mid \mathbf{y}^{\text{obs}})
\end{equation}

In practice, $p_{\text{ppc}}$ is approximated by simulation, through Algorithm \ref{alg:ppc}.

\begin{algorithm}[H]
        \caption{Posterior Predictive Check}
        \label{alg:ppc}
        \begin{algorithmic}[1]
        \STATE \textbf{Input:} Data $\mathbf{y}^{\text{obs}}$, posterior $p(\theta \mid \mathbf{y}^{\text{obs}})$, model $p(\mathbf{y} \mid \theta)$, and statistic $T(\mathbf{y}, \theta)$.        
        \FOR{$i = 1$ to $N$}
            \STATE Sample $\theta^{(i)} \sim p(\theta \mid \mathbf{y}_{\text{obs}})$.
            \STATE Sample replicated data $\mathbf{y}^{\text{rep},(i)} \sim p(\mathbf{y} \mid \theta^{(i)})$.
            \STATE Compute statistic for observed data: $T(\mathbf{y}^{\text{obs}}, \theta^{(i)})$.
            \STATE Compute stat. for replicated data: $T(\mathbf{y}^{\text{rep},(i)}, \theta^{(i)})$.
        \ENDFOR
        \STATE Calculate the posterior predictive $p$-value:
        \[
        p_{\text{ppc}} = \frac{1}{N} \sum_{i=1}^N \mathbb{I}\big(T(\mathbf{y}^{\text{rep},(i)}, \theta^{(i)}) \geq T(\mathbf{y}^{\text{obs}}, \theta^{(i)})\big)
        \]
        
        \STATE \textbf{Return:} $p_{\text{ppc}}$.
        \end{algorithmic}
\end{algorithm}

\subsubsection{PPC applied to choice models}

To test IIA with the PPC framework, we define a Bayesian version of the BTL model. For question sets $\overline{Q}_1, \ldots, \overline{Q}_m$, with targets $t_1, \ldots, t_m$, respectively,
\begin{align*}
    \sigma &\sim \text{HalfNorm}(2) \\
    s_{ik} &\sim \mathcal{N}(0, \sigma^2),\, i=1, \ldots, m, \,k \in T \setminus \{t_i\} \\
    a_{Q} &\sim \text{Mult}(n, \pi_Q(\mathbf{s}_i)), \, i=1, \ldots, m,\, Q \in \overline{Q}_i,
\end{align*}
that is, we define a half-normal hyperprior for the prior $\sigma$, and sample the similarity scores of items $k$ to target $t_i$ through a zero-mean Gaussian with standard deviation $\sigma$. Note that  $\sigma$ is shared across all question sets $\overline{Q}_i$, making this a hierarchical model (see graphical representation in Figure~\ref{fig:graph_iia} in Appendix \ref{sec:graphical_models}). 

When applied to survey data, the posterior distribution of $\sigma$ can be interpreted as the general magnitude of similarity scores. When $\sigma$ approaches 0, then most questions are answered close to uniformly at random. Conversely, if $\sigma$ is large, then one item is likely to stand out in each question. Having the generative model specified above, and having calculated the posteriors for all $s_{ik}$'s, we can the use the same statistic $D$ in Algorithm \ref{alg:ppc} to obtain a Bayesian version of the goodness of fit test.

\subsection{Population homogeneity}\label{sec:pop_homogeneity_theory}

A common assumption in the study of context effects and IIA is that of population homogeneity. In essence, participants are statistically equivalent in their similarity judgement of the questions they respond. It is also known that models like the Mixed Multinomial Logit model (MMNL) can violate IIA just by accounting for population heterogeneity \citep{mcfadden2000mixed, train2009discrete}. Thus, violations of IIA measured on real data can also be due to population heterogeneity, and not necessarily context effects induced by the choice-sets (see Appendix~\ref{app:heterogeneity} for a simple example). Thus, an additional step when quantifying IIA violations is assessing population homogeneity. If population is indeed homogeneous and IIA is violated, this provides stronger evidence of that relative similarity to the target depends on the choice-set. A statistical test based on PPC to measure population homogeneity is presented in Section~\ref{sec:pop_homogeneity}.

\section{Analysis of Synthetic Data}
\label{sec:synthetic}

Testing for IIA in similarity choices requires data where multiple questions share the same target. Moreover, the choice-sets of such questions must also overlap. In what follows we propose a model for generating synthetic data with such characteristics. This same data format will also be used in the user experiments to be presented.

\subsection{Generative model}
Let $Q^0$ denote a similarity question with a target $t_{Q^0}$ and choice-set $C_{Q^0} = \{c^1, c^2, c^3, c^4\}$. Question $Q^0$ is used to create four other similarity questions with the same target: $Q^{i}, i=1,\ldots,4$, where the choice-set $C_{Q^{i}} = C_{Q^0} \setminus \{ c_i \}$, namely dropping item $c_i$ once from the original choice-set. For example, the question $Q^{1}$ has choice-set $C_{Q^{1}} = \{c^2, c^3, c^4\}$. These five questions form a question set denoted by $\overline{Q}$. In what follows, three models are presented to generate $\pi_{Q^ik}$, namely the probability that item $c_k$ is chosen by a participant when presented question $Q^i$. 


{\bf IIA compliant.} Recall that under IIA, it is sufficient that every item has a similarity score to the target. Let $s_k \sim \mathcal{N}(0, \sigma^2)$ be a normally distributed and independent random variable for every $k=1,\ldots,4$. Given $s_k$, the following choice model is considered:
\begin{align}
    \pi_{Q^ik}(\mathbf{s}) = \frac{e^{s_k}}{\sum_{k' \in C_{Q^i}} e^{s_{k'}}}, \; k \in C_{Q^i}, \; i = 0,\ldots,4 
    \label{eq:IIA}
\end{align}
Note that every item $c_k$ is associated to a similarity score $s_k$, independently of the choice-set.

{\bf Additive perturbation to IIA.} In order to induce IIA violations, it is sufficient that the similarity scores of items to the target depend on the choice-set. The following model adds a perturbation to the baseline similarity scores. Let $\varepsilon^{i}_k \sim \mathcal{N}(0, \sigma_p^2)$ be a normally distributed and independent random variable for every $k=1,\ldots,4$ and $i=1,\ldots,4$. Note that the single parameter controlling $\varepsilon^{i}_k$ is $\sigma_p$. Conditioned on $\varepsilon^{i}_k$, the following choice model is considered:
\begin{align}
    \pi_{Q^ik}(\mathbf{s}) = \frac{e^{s_k + \varepsilon^{i}_k}}{\sum_{k' \in C_{Q^i}} e^{s_{k'}+\varepsilon^{i}_{k'}}}, \; k \in C_{Q^i}, \; i = 1,\ldots,4 
    \label{eq:additive}
\end{align}
Note that $Q^0$ is not perturbed. Moreover, if $\sigma_p = 0$, the additive perturbation becomes zero and the IIA compliant model is recovered; if $\sigma_p$ increases to large enough values the choice probabilities become relatively independent of each other. Thus, $\sigma_p$ is a parameter that controls how strong the additive model induces IIA violations. The additive perturbation model is a general description of IIA violations, without any particular mechanism for inducing context effects. In fact, even when an alternative perturbation model is used for generating data, fitting the additive perturbation model to the synthetic data results in an estimated positive $\sigma_p$  (see Appendix \ref{app:additive_sim}).

\subsection{Numerical evaluation}

Consider $m=100$ different question sets $\overline{Q}$, each generated independently by the generative models previously defined. Moreover, assume that each question in a question set is presented to $n=30$ simulated participants who all provide a simulated answer according to the choice probabilities defined by the respective model. 
Let $\sigma = 2$ for the IIA compliant model. Last, since $p$-values are random (since the dataset is random), the entire experiment is independently repeated 30 times, and the average of the minimum and aggregate $p$-values are presented.

\begin{figure}[t]
\centering
\includegraphics[width=1.0\linewidth]{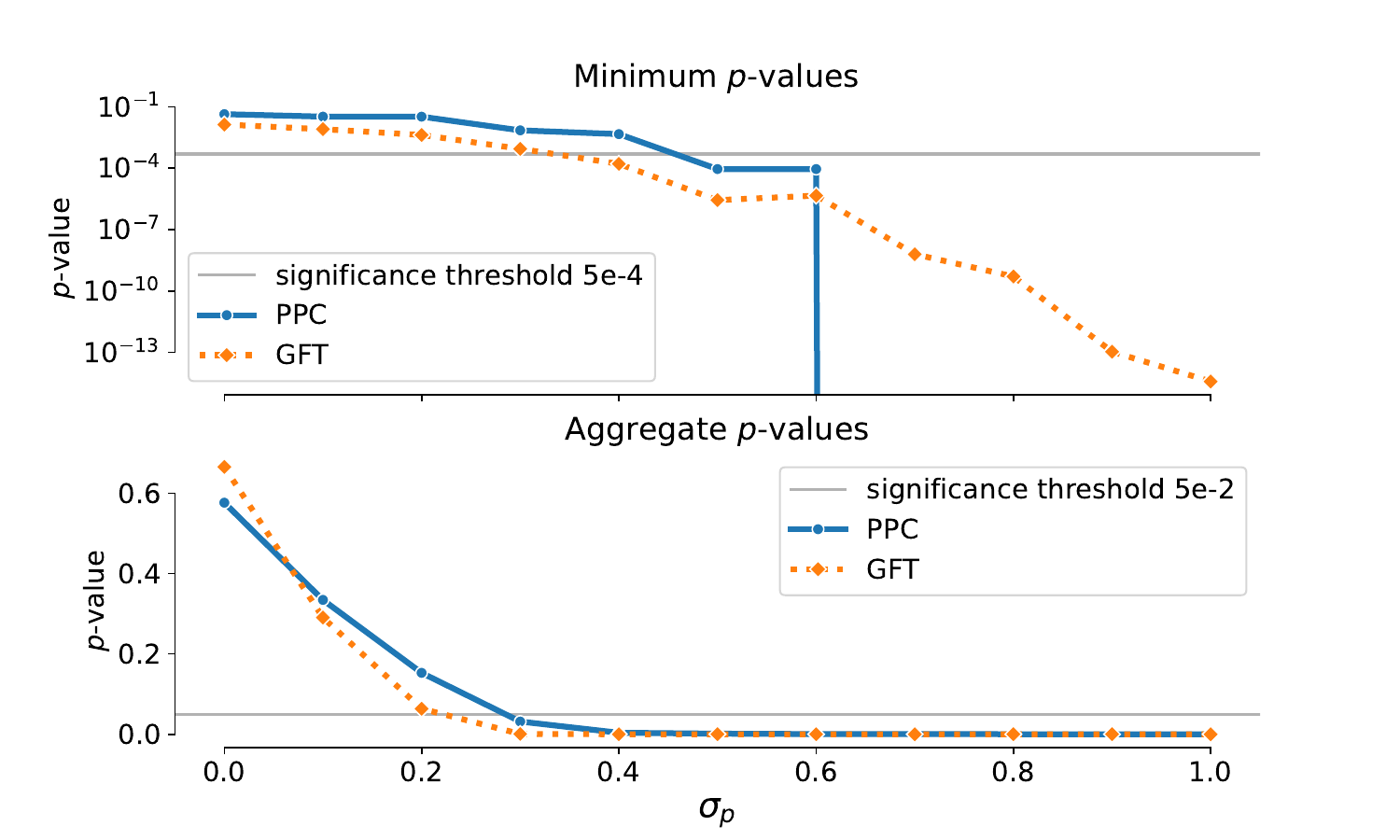}
\caption{$p$-values obtained by the statistical tests for IIA violations as a function of $\sigma_p$ for the additive perturbation model.}
\label{fig:test_additive}
\end{figure}

Figure~\ref{fig:test_additive} shows the $p$-values for data generated by the additive perturbation model as a function of $\sigma_p$, for both the minimum and aggregate $p$-values. Note that as $\sigma_p$ increases, the $p$-value for both statistical tests decreases, eventually cross the significance threshold of 0.0005 or 0.05 for the minimum and aggregate cases, respectively. However, the significance threshold for the minimum test requires a larger perturbation (around $\sigma_p = 0.35$ and $\sigma_p = 0.45$ for GFT and PPC, respectively) than in the aggregate test (around $\sigma_p = 0.2$ and $\sigma_p = 0.3$ for GFT and PPC, respectively). In essence, this is the amount of perturbation required for IIA to be rejected. Interestingly, the $p$-values for both tests decay relatively similar with $\sigma_p$ validating one another. Note that for PPC, when $\sigma_p \geq 0.6$ the $p$-value is zero since all samples from the posterior in the simulation where rejected. Last, an interesting phenomenon occurs at $\sigma_p = 0$ in the aggregate test; here, the $p$-value for the classical GFT is slightly larger than PPC, indicating that the null hypothesis is less likely to be rejected under GFT than PPC. 

\begin{figure}[t]
\centering
\includegraphics[width=1.0\linewidth]{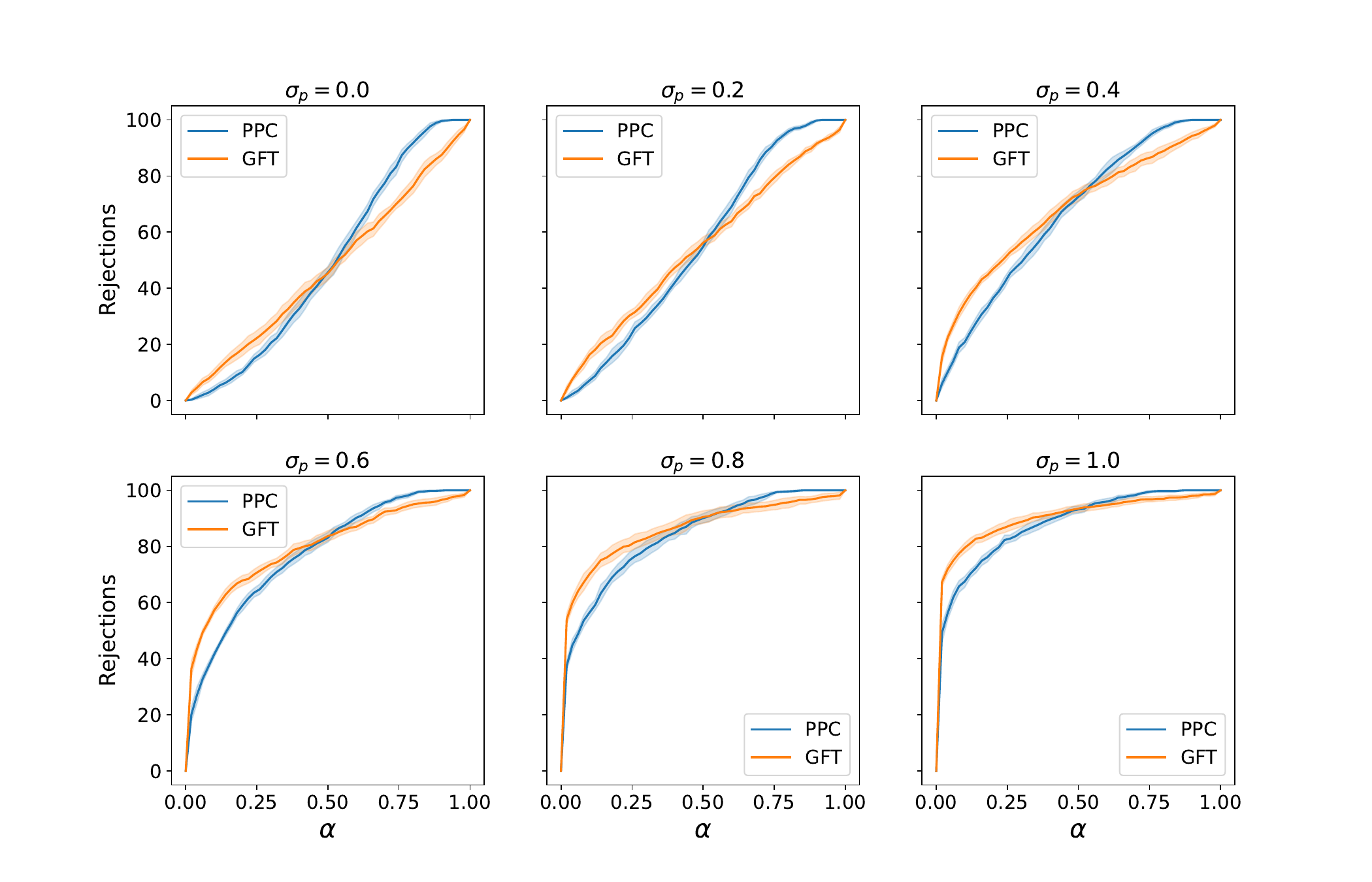}
\caption{Number of rejections as a function of the selection threshold $\alpha$ for different $\sigma_p$.}
\label{fig:rej_additive}
\end{figure}

The GFT and PPC tests can also be used to determine if a particular question set violates the null hypothesis. A selection threshold $\alpha$ can be applied to each question set, and question sets with a $p$-value below $\alpha$ are rejected under the null. Figure~\ref{fig:rej_additive} shows the number of rejections as a function of $\alpha$ for different $\sigma_p$ for both tests. Note that for $\sigma_p=0$, the number of rejections under GFT grows linearly with the threshold $\alpha$ as expected (under the null hypothesis, the $p$-values are uniformly distributed in the limit $n \rightarrow \infty$). However, PPC rejects less question sets for smaller values of $\alpha$. As $\sigma_p$ increases, GFT rejects more with very small $\alpha$ values and for $\sigma_p = 0.8$ around 60 question sets are rejected as soon as $\alpha$ is non-zero. PPC is slower to start rejections but it is faster to terminate rejecting all question sets. PPC rejects all question sets before $\alpha = 1$ while GFT requires $\alpha = 1$ to reject all. Thus, there is a tradeoff in these two statistical tests in the context of IIA.

\section{Experimental results}\label{sec:experiments}

In order to assess for IIA in similarity choices made by people, two different experiments have been designed in the form of web surveys. The items appearing in the questions to judge similarity surveys are images of dishes, fruits, snacks and food items in general. The set $T$ of 100 items used in the surveys was selected by manually curating the CROCUFID dataset \citep{CROCUFID}, so as to achieve the following properties:
\begin{enumerate}
    \item Variety: western and eastern food dishes, sweet and salty snacks, fruits, etc.
    \item Compositionality: items have single or multiple ingredients or combinations. For instance, meatball with mashed potatoes versus meatball alone.
    \item Perspective: the same item can appear from different visual perspectives. For instance, a whole loaf of bread versus bread slices. 
\end{enumerate}
Similarity judgement between the curated items can be drawn in many ways,  such as using ingredients, color, taste, and even culture associated with the items. Thus, the experiment serves as a prototypical setting for studying complex similarity judgements, and in particular, for testing for IIA. 

The web surveys designed have the same general structure: A participant provides her responses to 20 similarity questions; each question is comprised of one target food item displayed on the top of the screen and a choice-set displayed on the bottom (with three of four options); we ask ``Which option is most similar to the food item on top?'' to which the participant must respond by selecting exactly one item from the choice-set, before moving on to the next question (revising answers by returning to previous questions is also not allowed). The Prolific\footnote{Prolific is an online research platform with over 200k registered participants: \url{https://www.prolific.com/}} platform was used to solicit paid participants for the surveys, and no demographic filters were used when soliciting participants. On average, a participant completed the survey in 3 minutes\footnote{Excluding the time to log in the system and read the instructions.}. We provide as screenshot of the survey website in Appendix~\ref{sec:survey_website}.

\subsection{Handcrafted Dataset}

The first experiment is inspired by Tversky to show IIA violations and illustrate the {\em diagnosticity principle} in similarity judgements~\citep{tversky1977features}. In such experiment, questions are generated in pairs that have the same target and a single item difference in their choice-set. More precisely, $Q_a$ and $Q_b$ have the same target $t$ and choice-sets $C_{Q_a} = \{c_1, c_2, c_3\}$ and $C_{Q_a} = \{c_1, c_2, c_4\}$, respectively. Moreover, the questions should be designed such that $c_1$ and $c_2$ are comparatively more similar to $t$, and item $c_3$ or $c_4$ is more similar to $c_1$ or $c_2$, respectively, but also dissimilar from $t$. The general idea is that $c_3$ and $c_4$ change the context for the question, and can thus change the ratios of responses between $c_1$ and $c_2$ in the two questions (thus, violating IIA). 

\begin{figure}
    \centering
    \includegraphics[width=1.1\linewidth]{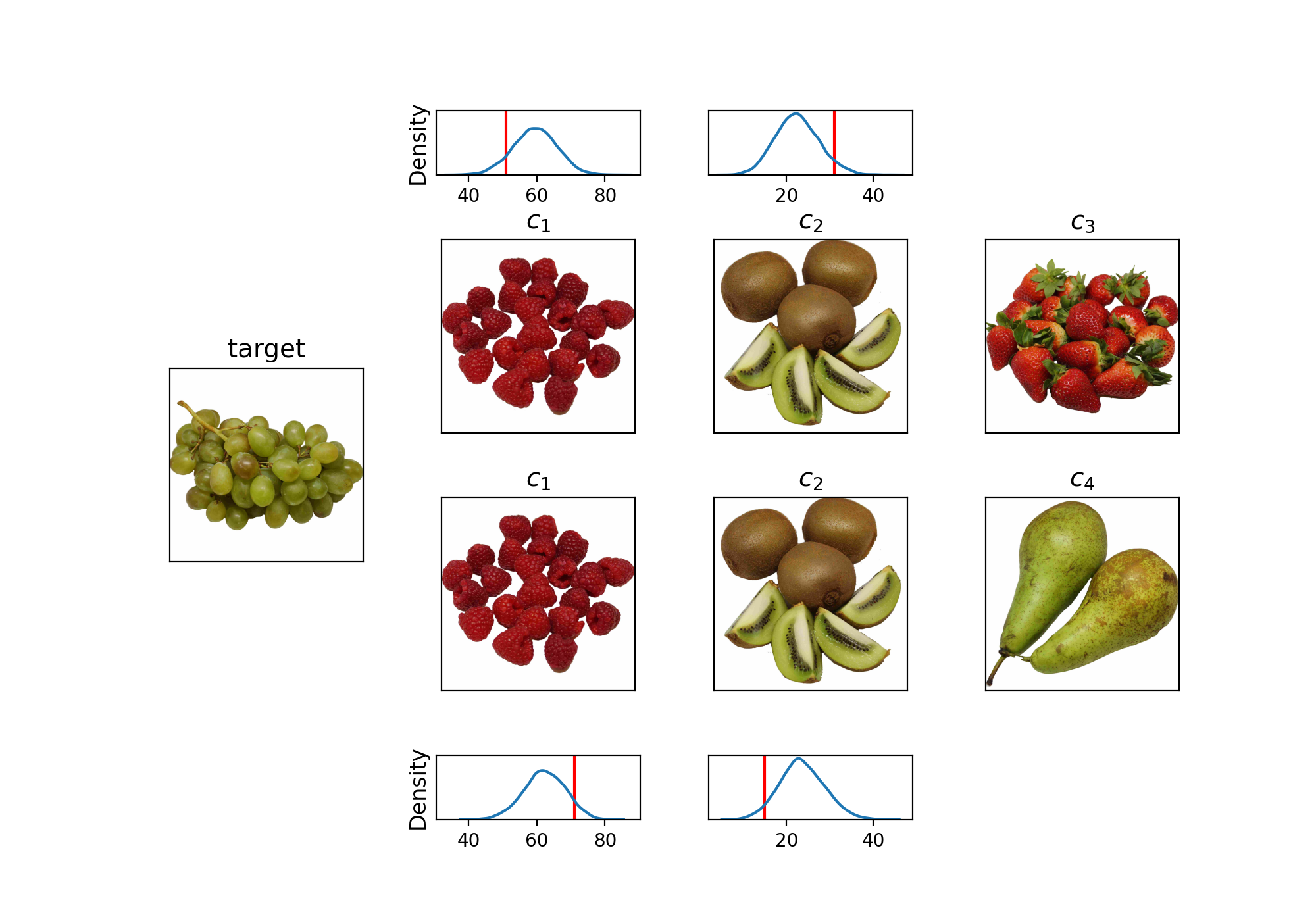}
    \caption{Example of a question pair from the survey. Vertical red line in the plots indicate number of participants selecting that item; blue curve shows the distribution of the (posterior) number of choices.}
    \label{fig:question_0058}
\end{figure}

A dataset consisting of 20 questions pairs ($Q_a, Q_b$) was manually built by the authors using the 100 curated food items. Each food item appears at most once in a survey (either as the target or in the choice-set) in order to minimize dependencies between question pairs. A participant in the survey was either presented with $Q_a$ or $Q_b$ but not both, for all 20 question pairs (see Appendix~\ref{sec:survey} for all twenty questions). The version $Q_a$ or $Q_b$ of a question pair was randomly chosen for each participant, as well as the order in which the participant answers the 20 questions\footnote{One question was used as a honey pot to flag spurious participants, and is not considered in the analysis.}. The total number of participants was 207.

Figure~\ref{fig:question_0058} illustrates the question pair with the smallest $p$-values for both GFT and PPC. Note that adding a red-coloured fruit (strawberry) in $Q_a$ increases the choices for the green fruit (kiwi), while adding a green-coloured fruit (pear) in $Q_b$ increases the choices of the red fruit (raspberry). This is a good example of what \cite{tversky1977features} calls the {\em diagnosticity principle} and a clear violation of IIA. Moreover, note that posterior distribution for the number of times an item is chosen in that choice-set under IIA is not a good model, given by the relative distance to the actual number of choices (red vertical bars).

The minimum $p$-values for GFT and PPC were 0.00052 and 0.0066, and thus the IIA is rejected by GFT ($\alpha=0.05/19 = 0.0026$). The aggregate $p$-values for GFT and PPC were 0.000015 and 0.041, and thus IIA is rejected by both tests. 

\begin{figure}
    \centering
    \includegraphics[width=1\linewidth]{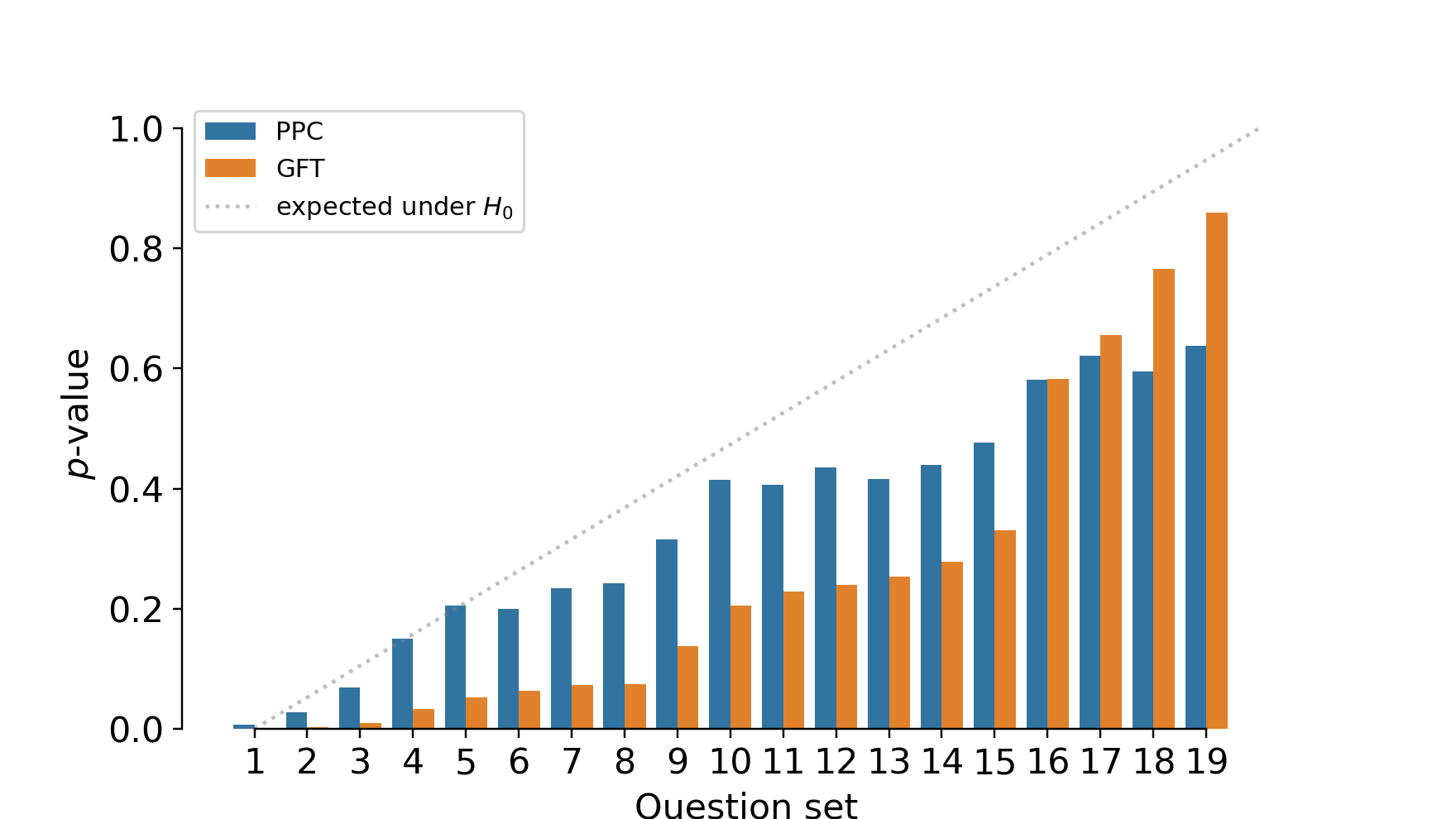}
    \caption{$p$-values obtained by PCC and GFT for each question set in the handcrafted dataset sorted by GFT value. Diagonal line corresponds to uniform distribution under IIA hypothesis.}
    \label{fig:handcrafted_pvalues}
\end{figure}

Figure~\ref{fig:handcrafted_pvalues} shows the $p$-values obtained for both tests for all questions in the survey (sorted by GFT). As with the synthetic dataset, GFT has smaller values than PPC. Note that under the joint null hypothesis (IIA), the $p$-values for GFT follow a uniform distribution, and thus the empirical CDF of $p$-value samples should follow a diagonal line, as indicated in the plot. The measured GFT $p$-values are below this diagonal line corroborating the rejection of the null hypothesis. 

\begin{figure}[h]
    \centering
    \includegraphics[width=1\linewidth]{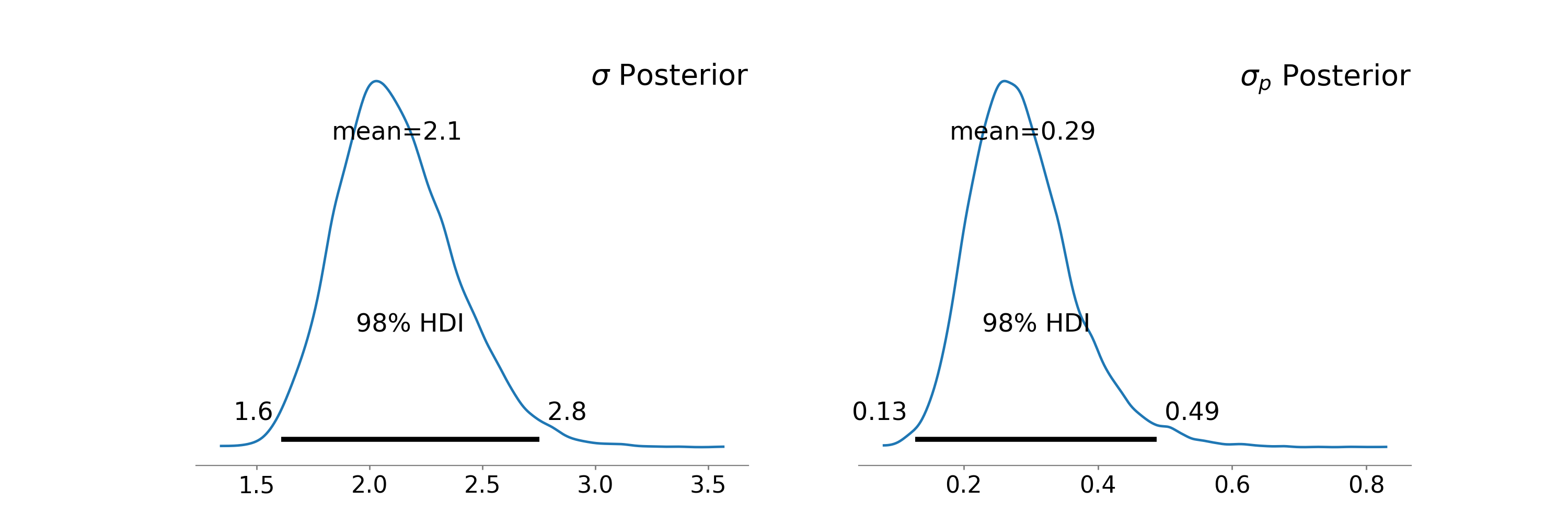}
    \caption{Posterior distributions for $\sigma$ and $\sigma_p$ given the handcrafted survey dataset. The mean values for $\sigma$ and $\sigma_p$ are 2.1 and 0.29, respectively.}
    \label{fig:additive_posterior_handcrafted}
\end{figure}

Besides testing for the IIA hypothesis, the additive perturbation model was also fitted to the handcrafted dataset. A $p$-value of 0.254 was obtained with PPC, thus implying this model can better represent this dataset (and not be rejected). Figure~\ref{fig:additive_posterior_handcrafted} shows the posterior distributions for $\sigma$ and $\sigma_p$ given the dataset. Interestingly, the posterior for $\sigma_p$ falls within $[0.12, 0.48]$ with high probability indicating it is an important component of the model. Moreover, the ratio between the average $\sigma_p$ and average $\sigma$ is $0.29/2.1 = 0.138$, indicating its relative magnitude is not insignificant. 

Interestingly, the average $\sigma_p$ when fitting the additive model to IIA compliant simulated data was 0.049, indicating this parameter plays a small role in this scenario (where IIA is present) but not on real data (see Figure~\ref{fig:additive_posterior_iia} in the Appendix). 

\subsection{Randomized Dataset}

The second experiment was designed to have a very different flavor. In contrast to manually curating question pairs, question sets were randomly determined using the curated food items. In particular, a total of 100 question sets were generated, each having a different target (thus, every food item in $T$ served as a target). For every target, a question $Q^0$ was generated by randomly selecting four food items for its choice-set. From $Q^0$, four questions were created by removing one of the items in its choice-set at a time, identical to the procedure described in Section~\ref{sec:synthetic}. Thus, $Q^i, i=1,\ldots,4$ have choice-sets with size 3.

While the total number of questions in this dataset is 500 (100 question sets each with 5 questions), the survey of a participant had only 20 questions, randomly chosen from the set of 500. However, in every participant survey, items in the target or choice-sets only appeared once. Last, every question received at least 18 responses, and 30 on average. 

\begin{figure}
    \centering
    \includegraphics[width=0.9\linewidth]{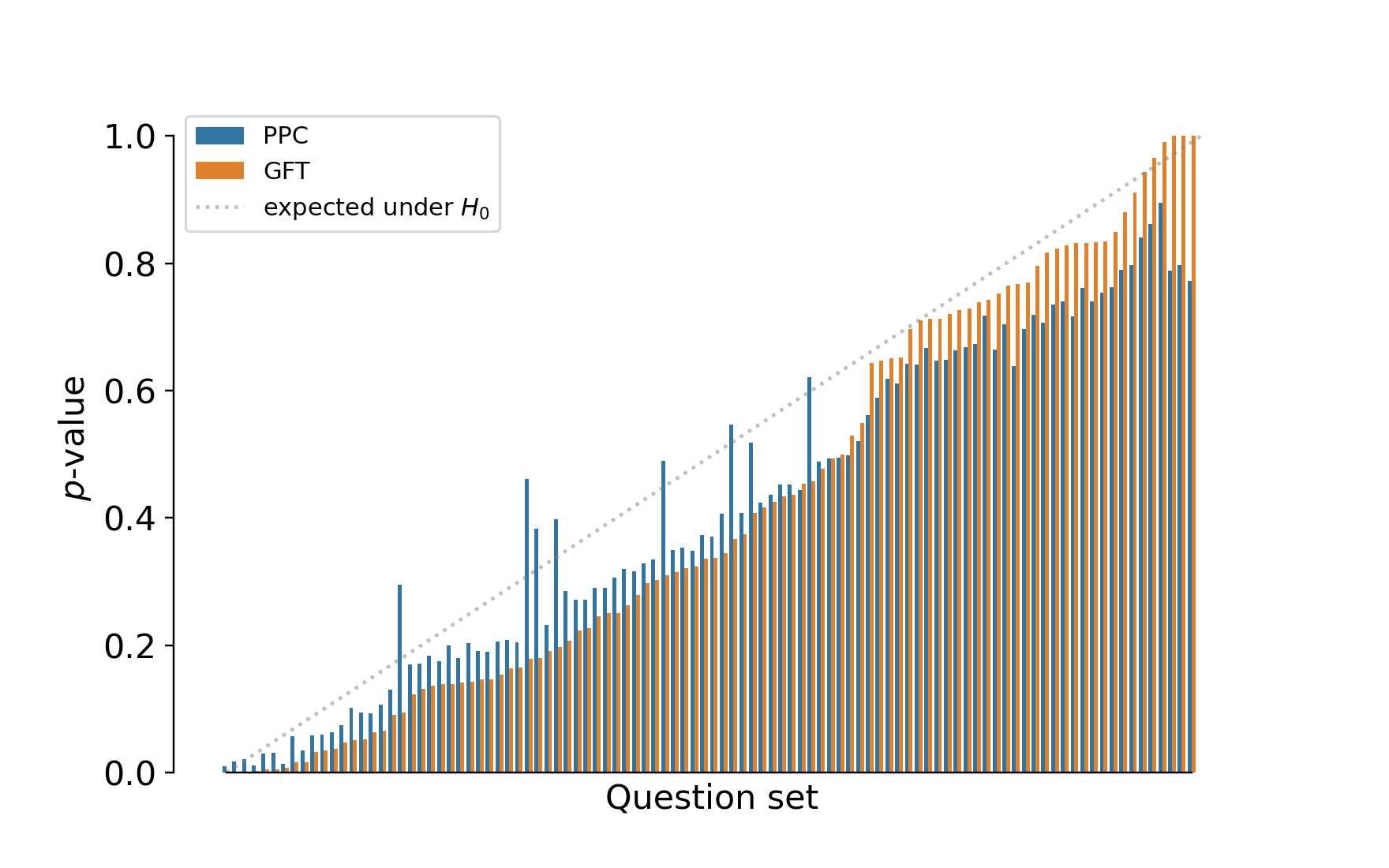}
    \caption{$p$-values obtained by PCC and GFT for each question set in the randomized dataset sorted by GFT value. Diagonal line corresponds to uniform distribution under IIA hypothesis.}
    \label{fig:random_pvalues}
\end{figure}

The minimum $p$-values for GFT and PPC were 0.0002 and 0.011, and thus the IIA is rejected by GFT ($\alpha=0.05/100 = 0.0005$). The aggregate $p$-values for GFT and PPC were 0.00002 and 0.0056, and thus IIA is rejected by both tests. The individual $p$-values per question set are shown in Figure \ref{fig:random_pvalues}.

\begin{figure}[h]
    \centering
    \includegraphics[width=1\linewidth]{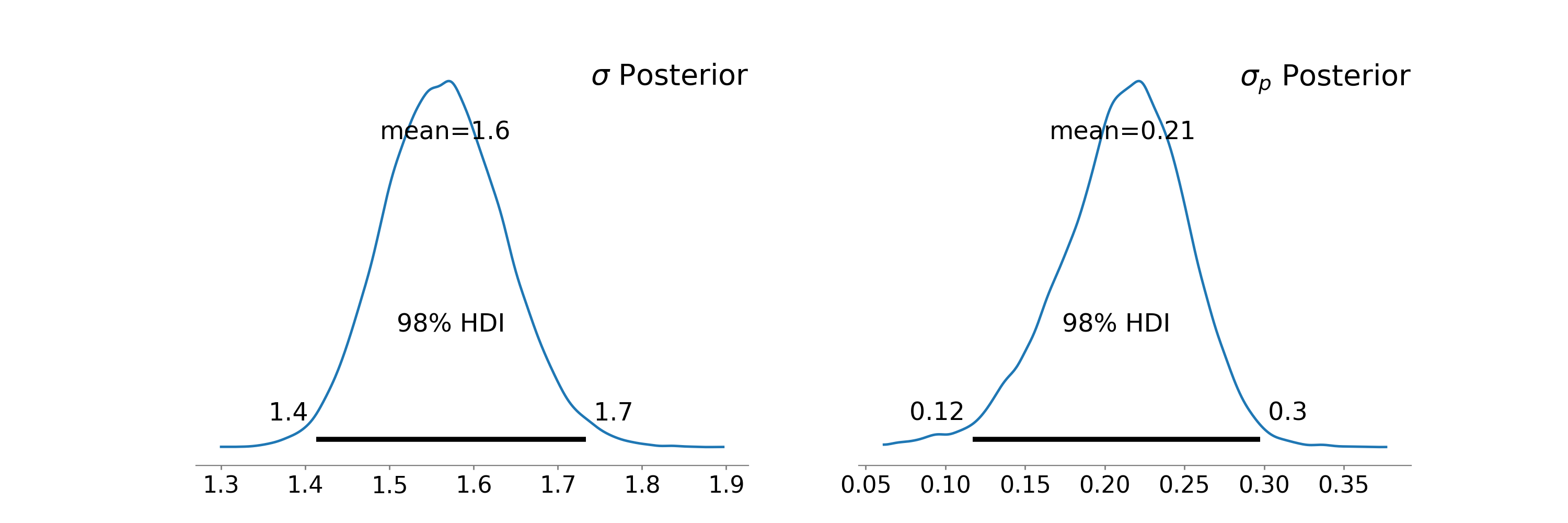}
    \caption{Posterior distributions for $\sigma$ and $\sigma_p$ after fitting the random survey data. The mean values for $\sigma$ and $\sigma_p$ are is 1.6 and 0.21, respectively.}
    \label{fig:additive_posterior_random}
\end{figure}

The additive perturbation model was also fitted to the randomized dataset and a $p$-value of 0.518 was obtained with PPC, implying this model can better represent this dataset (and not be rejected). Figure~\ref{fig:additive_posterior_random} shows the posterior distributions for $\sigma$ and $\sigma_p$ given this dataset. 
Interestingly, the posterior for $\sigma_p$ falls within $[0.14, 0.29]$ with high probability indicating its importance in fitting this model. The ratio between the average $\sigma_p$ and average $\sigma$ is $0.22/1.6 = 0.138$. Notably, this ratio is the same as for the handcrafted dataset. This result suggests that there is significant deviation from IIA even among randomly sampled similarity questions.

\section{Testing for population homogeneity}\label{sec:pop_homogeneity}

In this section, we develop a statistical test for population homogeneity (PH) based on the PPC framework. The motivation behind this test is to investigate whether a heterogenous population is a significant factor behind the observed IIA violations (see Section \ref{sec:pop_homogeneity_theory}). Our null hypothesis is that the respondents form homogenous population, as similarity comparisons are not too subjective (unlike preferences).

Suppose one has a survey of questions $\overline{Q}$. For each question $Q \in \overline{Q}$, one has a baseline distribution $\pi_Q$ specifying probabilities of responses for each question. Suppose a new participant takes the survey, and we want to test whether they follow the baseline distribution in their responses, or whether they display anomalous behaviour.
Let $I_p$ denote the {\em information content} (IC) of participant $p$, given by the negative log-probability of its selections, \textit{i.e.},
\begin{equation}
    I_p = - \sum_{Q \in \overline{Q}} \log \pi_{Qr_{pQ}}.
\end{equation}

A participant $p$ that answers according to the distribution $\pi_Q$, for all $Q \in \overline{Q}$ will have an $I_p$ whose expected value is the sum of entropies of each $\pi_Q$. A participant with a response distribution that is significantly different would have a much larger $I_p$. Therefore $I_p$ is an useful statistic to test whether a new participant $p$ follows the pre-specified parameters $\pi_Q$. The statistical test we propose is an extension of this basic idea to a set $P$ of participants and unknown parameters $\pi_Q$. We use the responses of the participants themselves to estimate the parameters, and then aggregate the $I_p$s of each participant into a single statistic, as we will see below.

Consider the experiment using the randomized dataset and a question set $\overline{Q}$ composed of all questions with four choices, $Q^0$. A total of 148 participants provide responses to these 100 questions and each participant answers 20 questions. Recall that each question has a unique target, and thus the similarity of items to the target can be treated independently (each item in each question as a similarity value). Thus, the MLE will simply be the empirical proportion of each chosen item $\pi_{Q_0k}(\hat{\mathbf{s}}) = n^{-1}a_{Q_0k}$, for $k \in C_{Q_0}$.

\begin{figure}[H]
    \centering
    \includegraphics[width=0.8\linewidth]{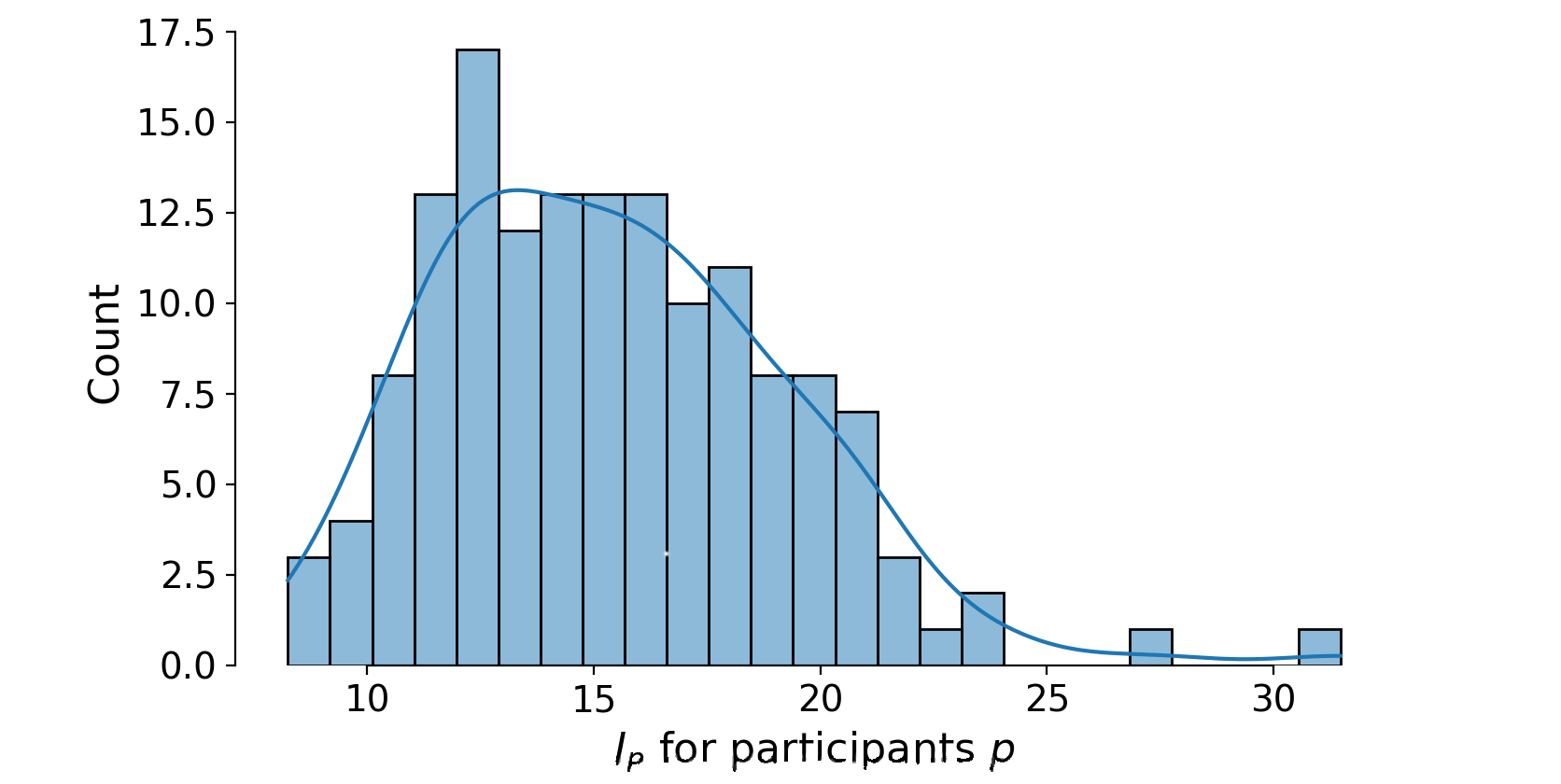}
    \caption{Distribution of the information content of participants.}
    \label{fig:IC}
\end{figure}

Figure~\ref{fig:IC} shows the distribution of the information content of the 148 participants. Note that the distribution appears skewed to the right, as indicated by two $I_p$'s higher than 25.

We define the test statistic for PPC to be the difference between the maximum and minimum information content among participants, namely
$$T = \max_{p \in P} I_p - \min_{p \in P} I_p.$$ We reject the null hypothesis (that the population is homogenous) if the $T$ is larger than what is expected under the null. We use the same PPC framework as before (Algorithm \ref{alg:ppc}).

PPC returns a $p$-value of 0.0315, thus rejecting PH. However, this is not surprising as the distribution of IC indicated the presence of an outlier. Moreover, when the responses of the single outlier participant is removed from the dataset, PPC returns a $p$-value is 0.27. Therefore, the null hypothesis cannot be rejected. This analysis suggests that the population is fairly homogeneous. Thus, the violations in IIA we observe are likely to stem from context effects.

\section{Summary and Future Work}
\label{sec:summary}

In this paper, we argue that it is important to test for the validity of the independence of irrelevant alternatives (IIA) property in similarity choice data. We also discuss that existing tests are not suitable for this purpose.
We propose two methods for this task, (1) the aggregation of goodness-of-fit statistics and (2) the application of posterior predictive checks (PPC) to a hierarchical Bayes model. Both tests give us a single $p$-value indicating the prevalence of IIA violations across the entire dataset. Moreover, an extension of the Bayesian model (the additive perturbation model) allows us to measure the strength of the IIA violations over the full dataset (see Figure \ref{fig:additive_posterior_handcrafted} and the accompanying discussion). Finally, we demonstrate the flexibility of the Bayesian model by using it to develop a test for population homogeneity. 

We apply these methods to similarity choice data that we collect through online anonymized surveys. The main findings of this paper indicate that IIA violations on similarity choice data are prevalent even under randomly generated questions. Indeed, this effect is as prominent in randomly generated questions as it is in handcrafted questions designed specifically to induce context effects. Further experiments confirm that population heterogeneity is not a factor causing these violations. Thus, our work provides convincing evidence that similarity choice data exhibits context effects. It motivates the development of richer choice models that can incorporate such effects,  perhaps by modelling known cognitive phenomena. In addition, this work also highlights the potential pitfalls of collecting similarity choice data with large choice-sets and breaking them down into triplets, as commonly done with the artist similarity dataset \citep{ellis2002quest} and the food similarity dataset \citep{wilber2014cost}.

\begin{acknowledgements} 
This work received financial support through research grants from CNPq
(402689/2019-4 and 310742/2023-4) and FAPERJ (E-26/200.483/2023), and from the Swiss National Science Foundation (SNSF) under grant IZBRZ2-186313.
\end{acknowledgements}

\bibliography{refs}

\newpage 

\onecolumn

\title{Measuring IIA Violations in Similarity Choices with Bayesian Models\\(Supplementary Material)}
\maketitle


\appendix

\section{IIA implies model with at most $T-1$ parameters}
\label{app:IIA}

The IIA assumption implies that $\pi_Q$, for all questions $Q$ having the same target can be fully specified with at most $|T|-1$ parameters, one per item in the set $T$ excluding the target. To see this is true, let $Q^*$ be a question with a choice-set that has all items except its target $t_{Q^*} = \ell$, thus $C_{Q^*} = T \setminus \{ \ell \}$. For any question $Q$ with the same target $\ell$, we have that 
$$ \frac{\pi_{Qk}}{\pi_{Q^*k}} = \frac{\pi_{Qk'}}{\pi_{Q^*k'}}, \, \forall k, k' \in C_{Q}. $$
These equalities across all item pairs $k, k' \in C_{Q}$ imply that the response probabilities in $Q^*$ provide the response probabilities for questions $Q$ as follows
$$ \pi_{Qk} = \frac{\pi_{Q^*k}}{\sum_{l\in C_Q} \pi_{Q^*l}}. $$

Thus, assuming that IIA holds, it is sufficient to specify a similarity score $s_k$ for every item $k \in T \setminus \{\ell\}$ to a fixed target $\ell$, independent of the question $Q$. For instance, on can take $s_k = \log \pi_{Q^*k}$. Therefore, without loss of generality, the response probability can be represented by the following model parametrized by the similarity vector $\mathbf{s}$:
\begin{equation}
\pi_{Qk}(\mathbf{s}) = \frac{e^{s_k}}{\sum_{k' \in C_Q} e^{s_{k'}}} \; , 
\label{eq:pi_s2}
\end{equation}
specifying that the probability of choosing item $k$ from choice-set $C_Q$ is proportional to $e^{s_k}$. 

\section{Violating IIA with population heterogeneity}
\label{app:heterogeneity}

In order illustrate how a mixture of two populations can violate IIA, consider a question set with two questions and the following choice-sets: $Q_{C_1} = \{a, b, c\}$, and $Q_{C_2} = \{a, b, d\}$. Consider two participant populations $p_1$ and $p_2$, each of them homogeneous but with different preferences, as shown in Table~\ref{tab:heterogeneous}. 

\begin{table}
        \centering
        \begin{tabular}{c|c|c|c}
             & $p_1$ & $p_2$ & mixt. $p_1$ $+$ $p_2$\\
             \hline 
             $(a, b, c)$ & 0.4, 0.6, 0  & 0.09, 0.01, 0.9 & 0.25, 0.3, 0.45 \\
             $(a, b, d)$ & 0.2, 0.3, 0.5 & 0.9,  0.1,  0  & 0.55, 0.2, 0.25 \\
        \end{tabular}
        \caption{IIA violation example, due just to population heterogeneity.}
        \label{tab:heterogeneous}
\end{table}
Note that for both $p_1$ and $p_2$ the odd ratios between items $a$ and $b$ are invariant in the two questions (2:3 and 9:1, respectively). Thus, each population conforms to IIA. However, under an equal mixture of $p_1$ and $p_2$ (i.e., 50\% each), the response probabilities for each question will change, as shown in Table~\ref{tab:heterogeneous}. In the mixed population, the ratios between items $a$ and $b$ in the two questions are no longer equal, violating IIA. 

\section{Numerical methods}
\label{app:numerical_methods}

We used the PyMC (\cite{AbrilPla2023}) to implement our bayesian models and estimate the model posteriors by MCMC sampling, using the NUTS algorithm (\cite{NUTS}). 

For executing the goodness of fit test with the $\chi^2$ statistic, we obtained the MLE executing a simple gradient descent algorithm. We used a learning rate of 0.005, and a stopping criterion of a less than $10^{-4}$ improvement in the log-likelihood of the parameters. If for a given question set $\overline{Q}$ with the same target $t$, some item $k$ was never selected, i.e. $a_{Qk} = 0, \forall Q \in \overline{Q}$, then we excluded item $k$ as an option from the data. That way, we have a bounded optimization problem without the need for regularization, which was not used. 

\section{Graphical model diagrams}\label{sec:graphical_models}

First, we show in Figure \ref{fig:graph_iia} the plate notation for the Bayesian BTL model. The standard deviation $\sigma$ of similarity scores $s$ is sampled from a half-normal distribution with parameter $\alpha_\sigma$. For each target $t_i, i=1, \ldots, m$, we sample $s_{ik}$ from the normal with $\sigma$. For a given question $Q$ with target $t_i$, The number of participants that selected option $k$ will be sampled from a multinomial distribution parametrized by the softmax of all similarities $s_{ik}$ with $k \in C_Q$.

\begin{figure}[h]
    \centering
    \begin{tikzpicture}

        \node[latent] (sigma) {$\sigma$};
        \node[latent, below=1.4cm of sigma] (s) {$s_{ik}$};
        \node[latent, right=2.3cm of s] (a) {$a_Q$};
        \node[const, above=1.2cm of sigma] (asigma) {$\alpha_\sigma$};

        \factor[above=0.6cm of sigma]     {sigma-f}     {right:HalfNormal} {} {} ; %
        \factor[above=0.8cm of s]     {s-f}     {right:Normal} {} {} ; %
        \factor[left=1.2cm of a]     {a-f}     {above:Multinomial} {} {} ; %

        \factoredge {asigma} {sigma-f} {sigma} ; %
        \factoredge {sigma} {s-f} {s};
        \edge {s} {a};

        \plate {plateK} {(s)} {$k \in T \setminus \{t_i\}$} ;
        \plate {plateQ} {(a)} {$Q \in \overline{Q}_i$};
        \plate[inner sep=.1cm] {plateI} {(plateQ) (plateK)} {$i=1,\dots,m$};

    \end{tikzpicture}
    \caption{Graphical model representation for the IIA model}
    \label{fig:graph_iia}
\end{figure}
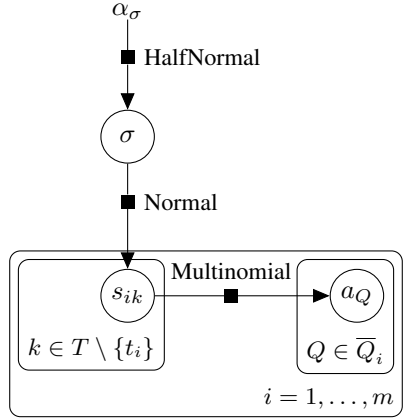

\begin{figure}[h]
    \centering
    \begin{tikzpicture}

        \node[latent] (sigma) {$\sigma$};
        \node[latent, right=3.5 cm of sigma] (sigma_p) {$\sigma_p$};
        \node[latent, below=1.4cm of sigma] (s) {$s_{ik}$};
        \node[latent, below=1.4cm of sigma_p] (p) {$\varepsilon_{Qk}$};
        \node[latent, below=1.5cm of a-f] (a) {$a_Q$};
        \node[const, above=1.2cm of sigma] (asigma) {$\alpha_\sigma$};
        \node[const, above=1.2cm of sigma_p] (bsigma) {$\beta_\sigma$};

        \factor[above=0.6cm of sigma]     {sigma-f}     {right:HalfNormal} {} {} ; %
        \factor[above=0.8cm of s]     {s-f}     {right:Normal} {} {} ; %
        \factor[above=0.6cm of sigma_p]     {sigmap-f}     {right:HalfNormal} {} {} ; %
        \factor[above=0.8cm of p]     {sp-f}     {right:Normal} {} {} ; %
        \factor[right=0.88cm of s]     {a-f}     {above:Multinomial} {} {} ; %

        \factoredge {asigma} {sigma-f} {sigma} ; %
        \factoredge {sigma} {s-f} {s};
        \factoredge {bsigma} {sigmap-f} {sigma_p} ; %
        \factoredge {sigma_p} {sp-f} {p};
        \factoredge {s} {a-f} {a};
        \factoredge {p} {a-f} {a};

        \plate {plateK} {(s)} {$k \in T \setminus \{t_i\}$} ;
        \plate {plateP} {(p)} {$Q \in \overline{Q}_i, k \in C_Q$}
        \plate[inner sep=.1cm] {plateQ} {(a)} {$Q \in \overline{Q}_i$};
        \plate[inner sep=.1cm] {plateI} {(plateQ) (plateK) (plateP)} {$i=1,\dots,m$};

    \end{tikzpicture}
    \caption{Graphical model representation for the additive perturbation model}
    \label{fig:graph_add_pert}
\end{figure}
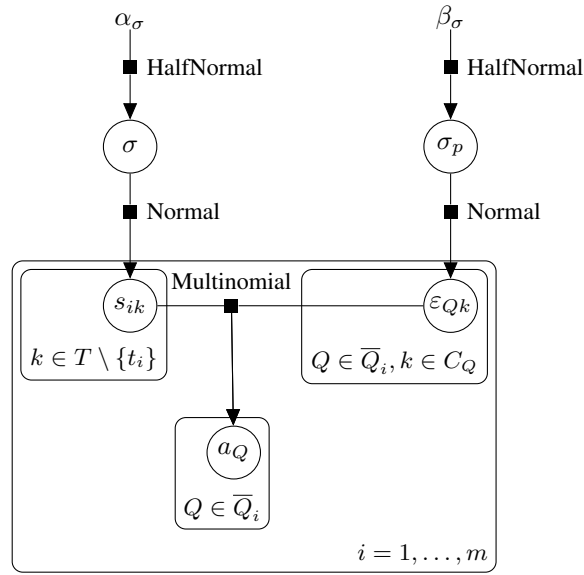

In Figure \ref{fig:graph_add_pert}, we show the plate notation for the additive perturbation model. In addition to $\sigma$ and the subsequent $s_{ik}$'s, we also have per question/item perturbation terms $\varepsilon$. Similar to $s_{ik}$, the noises $\varepsilon_{Qk}$ are sampled from a normal distribution, whose standard deviation $\sigma_p$ is sampled from a half-normal hyper-prior controlled by $\beta_\sigma$. For a given target $t_i$, all questions containing a certain item $k$ will attribute to it the same similarity score $s_{ik}$, but every question $Q \in Q_i$ will have a distinct perturbation term $\varepsilon_{Qk}$ added to that similarity. The perturbed similarities will then be put through a softmax to determine the parameters of the multinomial distribution generating outcomes $a_Q$.

\section{Model Posteriors}

\subsection{Additive perturbation model applied to simulated datasets}
\label{app:additive_sim}

\begin{figure}[h]
    \centering
    \includegraphics[width=0.9\linewidth]{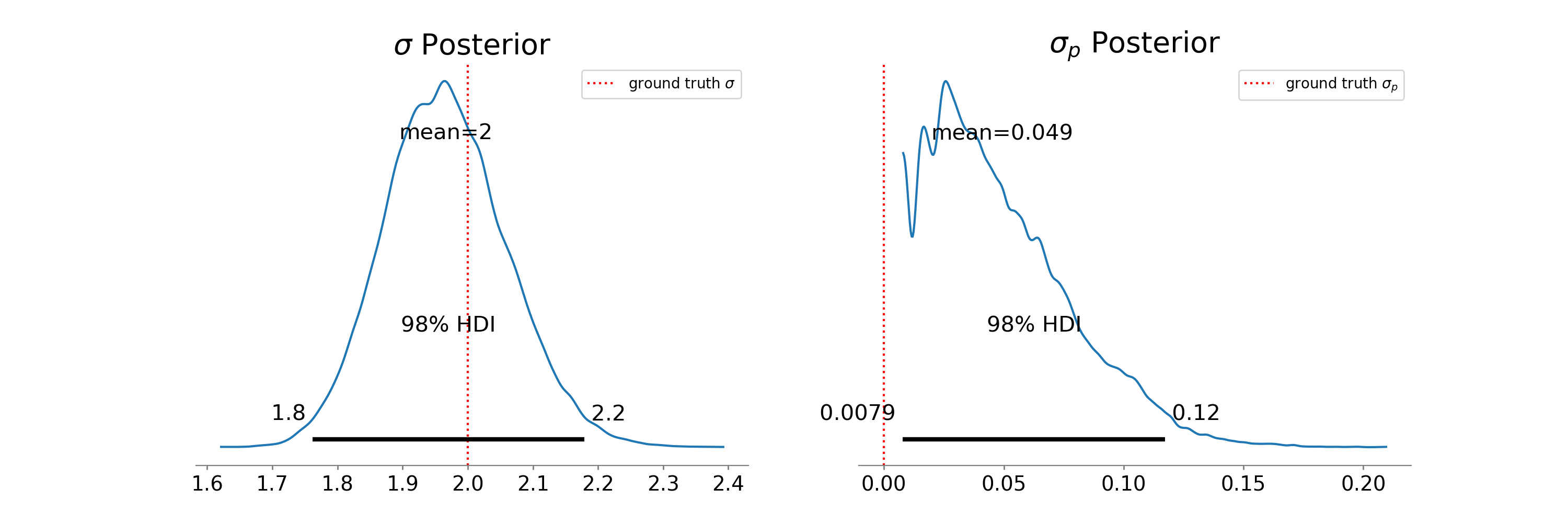}
    \caption{Posterior distributions for $\sigma$ and $\sigma_p$ after fitting the additive perturbation model to the IIA compliant simulated data.}
    \label{fig:additive_posterior_iia}
\end{figure}

We fitted the additive perturbation model to IIA-compliant simulated data with ground truth $\sigma=2$. We set the $\sigma$ hyper-prior parameter at $\alpha_\sigma = 1.5$ and the $\sigma_p$ hyper-prior parameter at $\beta_\sigma = 1$. The posterior distribution was estimated by executing the NUTS algorithm with 4 chains and 40000 samples each (burn-in of 20000). By applying PPC, we obtained a $p$-value of 0.5001, thus implying the model to be a good fit to the data. Moreover, the posterior $\sigma_p$ average was 0.058, while the ground-truth of $\sigma$ was recovered. This simulation shows that the additive perturbation model is well-behaved and identifies the lack IIA violations. See Figure~\ref{fig:additive_posterior_iia}. 

\begin{figure}[h]
    \centering
    \includegraphics[width=0.9\linewidth]{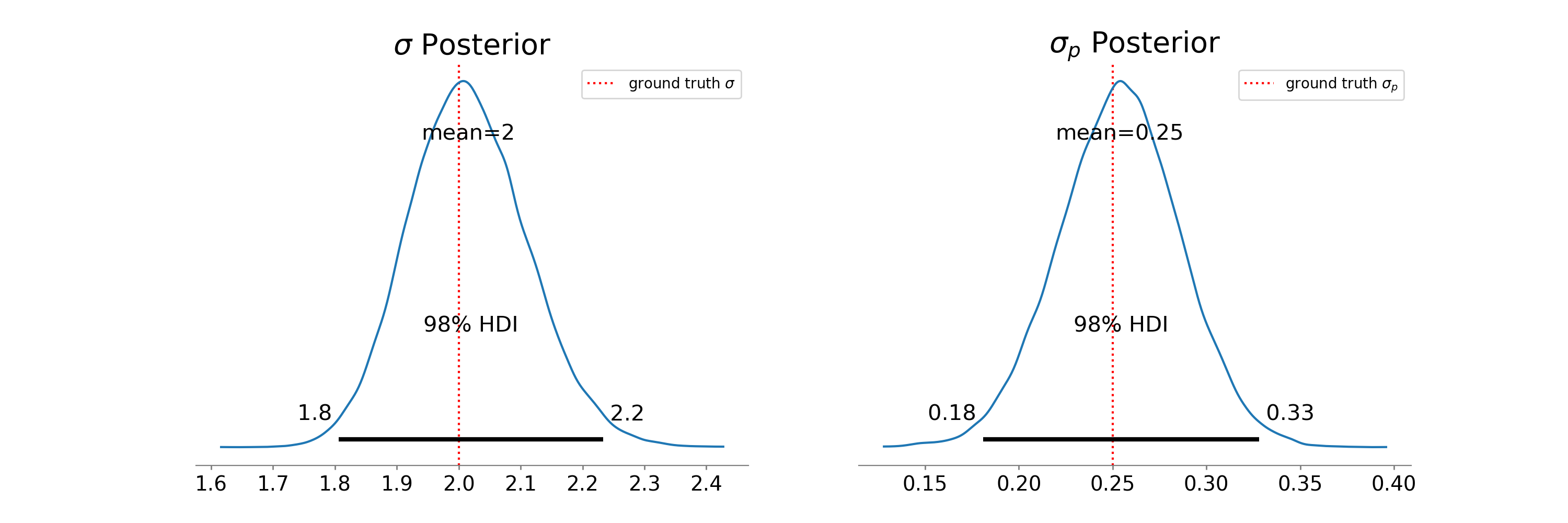}
    \caption{Posterior distributions for $\sigma$ and $\sigma_p$ after fitting the additive perturbation model to the additive perturbation simulated data.}
    \label{fig:additive_posterior_add_sim}
\end{figure}

We fitted the additive perturbation model to IIA-violating (from additive perturbation) simulated data, with ground truths $\sigma = 2$ and $\sigma_p = 0.2$. Again, we set the $\sigma$ hyper-prior parameter at $\alpha_\sigma = 1.5$ and the $\sigma_p$ hyper-prior parameter at $\beta_\sigma = 1$. The posterior distribution was estimated by executing the NUTS algorithm with 4 chains and 20000 samples each (burn-in of 10000). Through PPC we obtained a $p$-value of 0.6, thus implying the model to be a good fit to the data, unsurprisingly. Moreover, the posterior averages of $\sigma$ and $\sigma_p$ matched the ground truths. See Figure~\ref{fig:additive_posterior_add_sim}.

We also fitted the additive perturbation model to data generated with the multiplicative perturbation model (defined in Appendix ~\ref{app:multiplicative}) and found that when $\sigma_m$ is high, we get an estimated positive $\sigma_p$. More precisely, we generated 100 questions, with 30 responses each, following the logic described in Section ~\ref{sec:synthetic}. With $\sigma_m = 0.1$, the posterior distribution of $\sigma_p$ had a 2.5th percentile of 0.027 (averaged over 10 runs), which is close to 0, however, when we increased $\sigma_m$ to 0.2 and 0.3, we obtained 2.5th percentiles of 0.08 and 0.24, respectively, which are more distant from 0. In all cases,  $\sigma=2$ was used.

\section{Multiplicative perturbation model}
\label{app:multiplicative}

{\bf Multiplicative perturbation to IIA.} Similar to the additive above, this model also adds perturbations to the the original similarity scores. However, it does so using a single noise parameter per question in a multiplicative fashion. Thus, it is a simpler alternative model to induce IIA violations. Let $\varepsilon^{i} \sim \mathcal{N}(1, \sigma_m)$ be a normally distributed and independent random variable for every $i=1,\ldots,4$. Assuming $\varepsilon^{i}$, the following Bayesian choice model is considered:
\begin{align}
    \pi_{Q^ik}(\mathbf{s}) = \frac{e^{s_k \varepsilon^{i}}}{\sum_{k' \in C_{Q^i}} e^{s_{k'} \varepsilon^{i}}}, \; k \in C_{Q^i}, \; i = 1,\ldots,4 
    \label{eq:multiplicative}
\end{align}
Note that $Q^0$ is not perturbed. Moreover, if $\sigma_m = 0$, the multiplicative perturbation variable becomes one and the IIA compliant model is recovered; note that a large and positive $\varepsilon^{i}$ will magnify the similarity score differences, and thus violate IIA, but will preserve the preference ordering among the choice-set; a negative $\varepsilon^{i}$ will invert the ordering. Thus, $\sigma_m$ is a knob that controls how strong the multiplicative perturbation model induces IIA violations. Last, note that the two perturbation models (additive and multiplicative) are relatively different in their mechanism to induce IIA violations. 

\begin{figure}[h]
\centering
\includegraphics[width=0.55\linewidth]{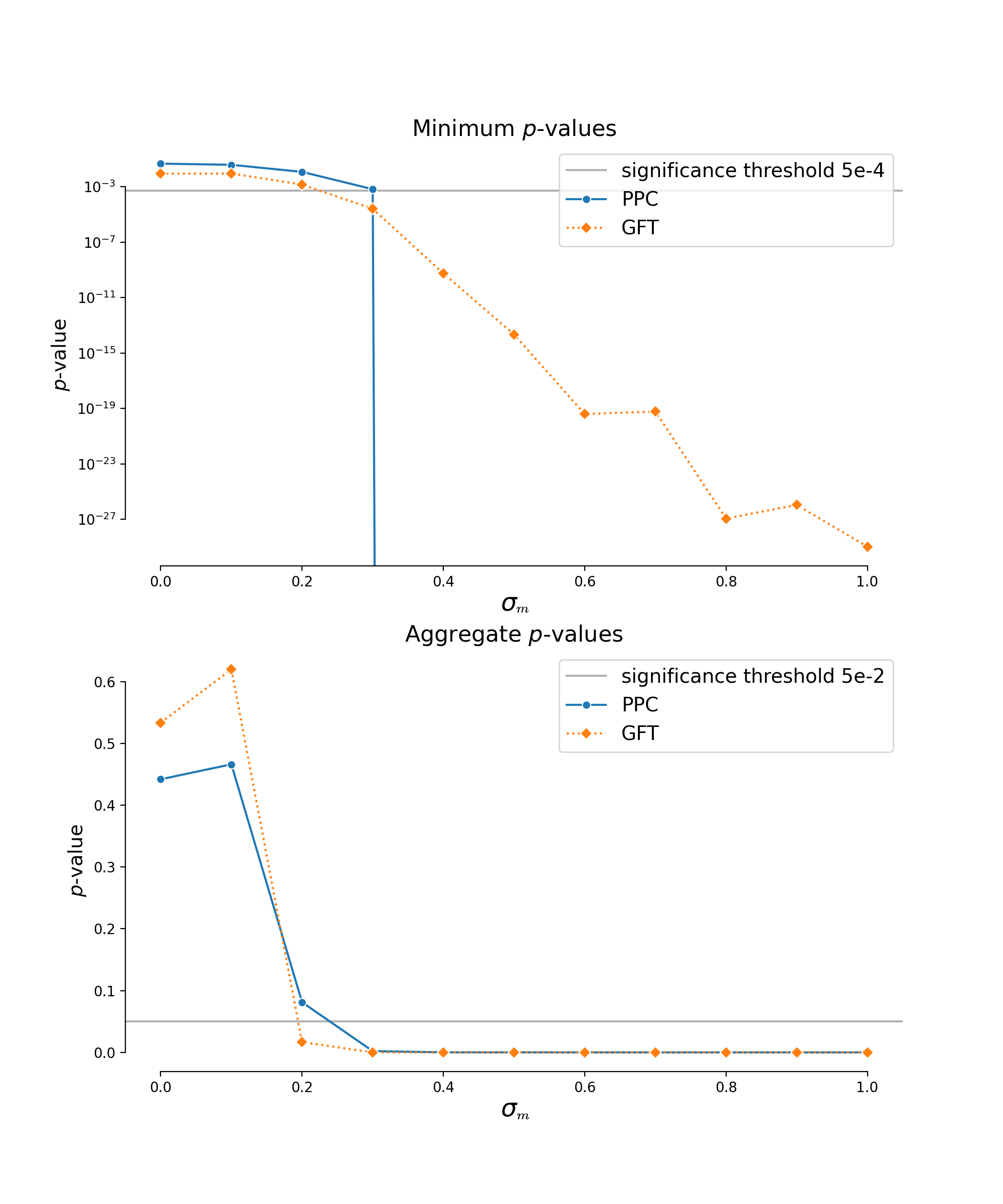}
\caption{$p$-values obtained by the statistical tests for IIA violations as a function of $\sigma_p$ for the multiplicative perturbation model.}
\label{fig:test_multiplicative}
\end{figure}

Figure~\ref{fig:test_multiplicative} shows the $p$-values for the multiplicative perturbation model as a function of $\sigma_p$, for both the minimum and aggregate $p$-values. Again, note that that as $\sigma_p$ increases the $p$-values decrease, eventually crossing the significance threshold. Interestingly, for both minimum and aggregate cases, a smaller value for $\sigma_p$ is required to cross the significance threshold, in comparison to the additive model (see Fig.~\ref{fig:test_additive}). This suggests that the multiplicative model introduces stronger violations of IIA for the same $\sigma_p$ (although the two models are not directly comparable). Again, both GFT and PPC behave relatively similar in both cases. 

\subsection{Multiplicative perturbation model applied to randomized dataset}

\begin{figure}[h]
    \centering
    \includegraphics[width=0.9\linewidth]{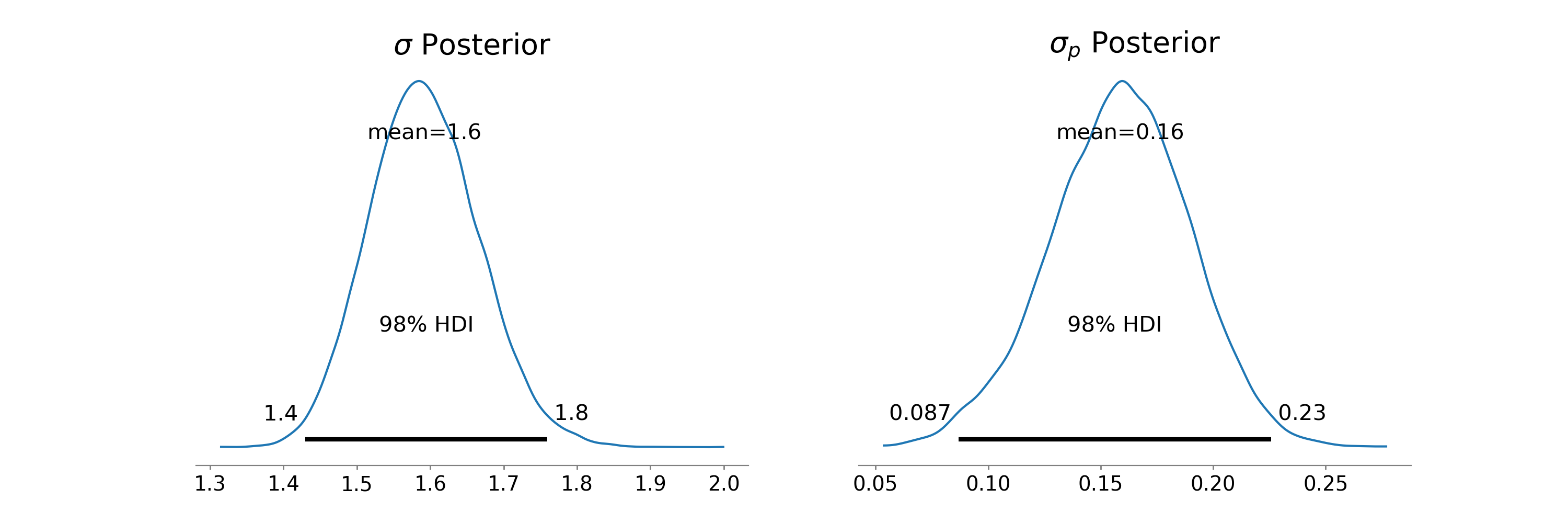}
    \caption{Posterior distributions for $\sigma$ and $\sigma_p$ after fitting the randomized dataset to the multiplicative pertubation model.}
    \label{fig:multiplicative_posterior_random}
\end{figure}

We fitted the multiplicative perturbation model to the randomized survey dataset, and obtained a $p$-value of 0.066 with PPC, failing to reject the model. The $p$-value is however, low enough for us to infer that the multiplicative model is unlikely to explain the range of context effects in the data. Figure~\ref{fig:multiplicative_posterior_random} shows the posterior distribution for both $\sigma$ and $\sigma_p$. Note that their mean values are $1.6$ and  $0.16$, respectively, indicating that $\sigma_p$ contributes to explaining the dataset, as is the case for the additive perturbation model. 

\section{The survey website}\label{sec:survey_website}

\begin{figure}[H]
    \centering
    \includegraphics[width=0.5\linewidth]{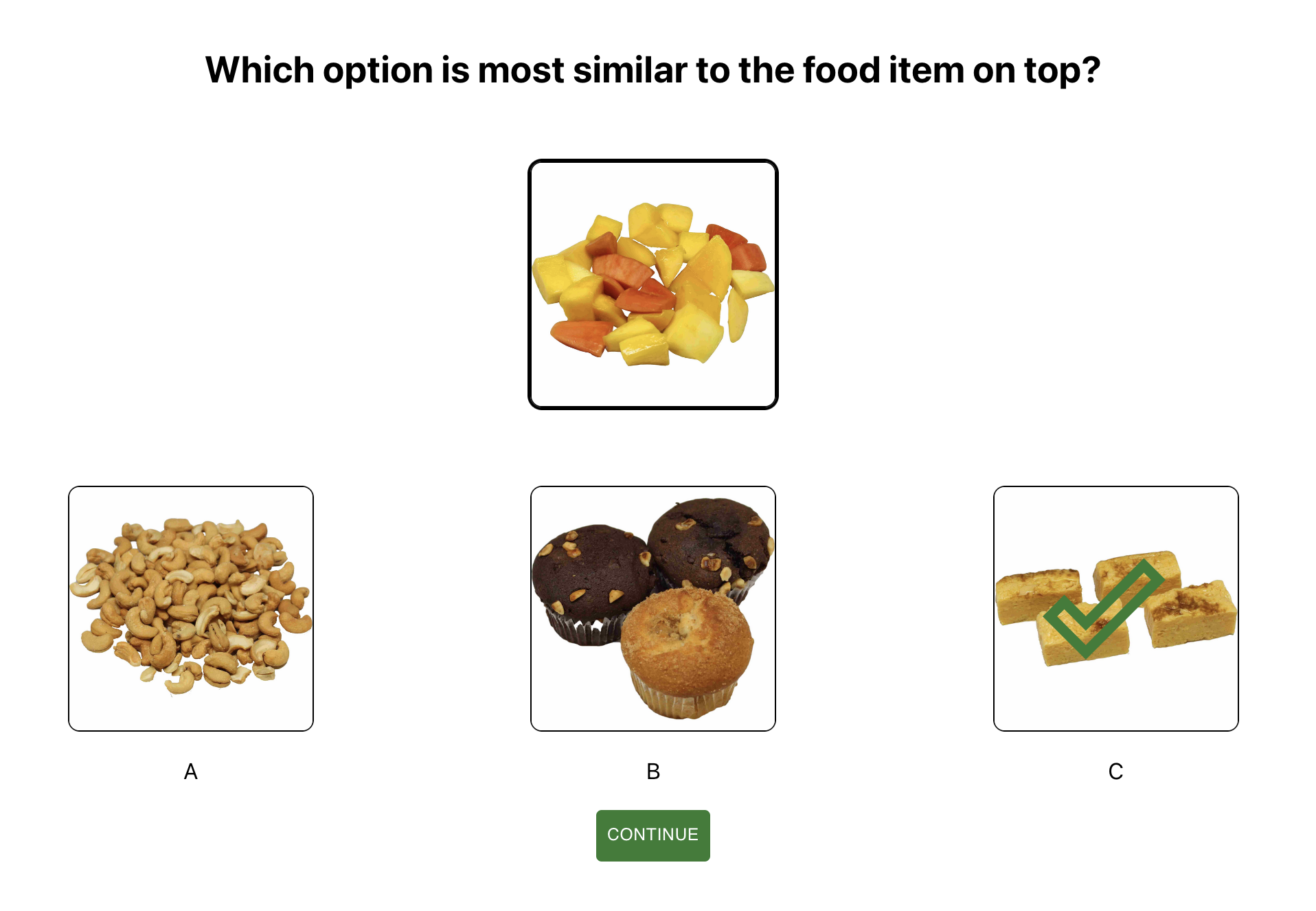}
    \caption{Screenshot of a typical survey question asked to participants on Prolific.}
    \label{fig:survey_screenshot}
\end{figure}

\section{Question pairs in the handcrafted dataset}\label{sec:survey}
The following figures, Combined with Figure \ref{fig:question_0058}, display the question pairs (and the response statistics) that were asked in the handcrafted dataset.
\begin{figure}
    \centering
    \includegraphics[width=0.65\linewidth]{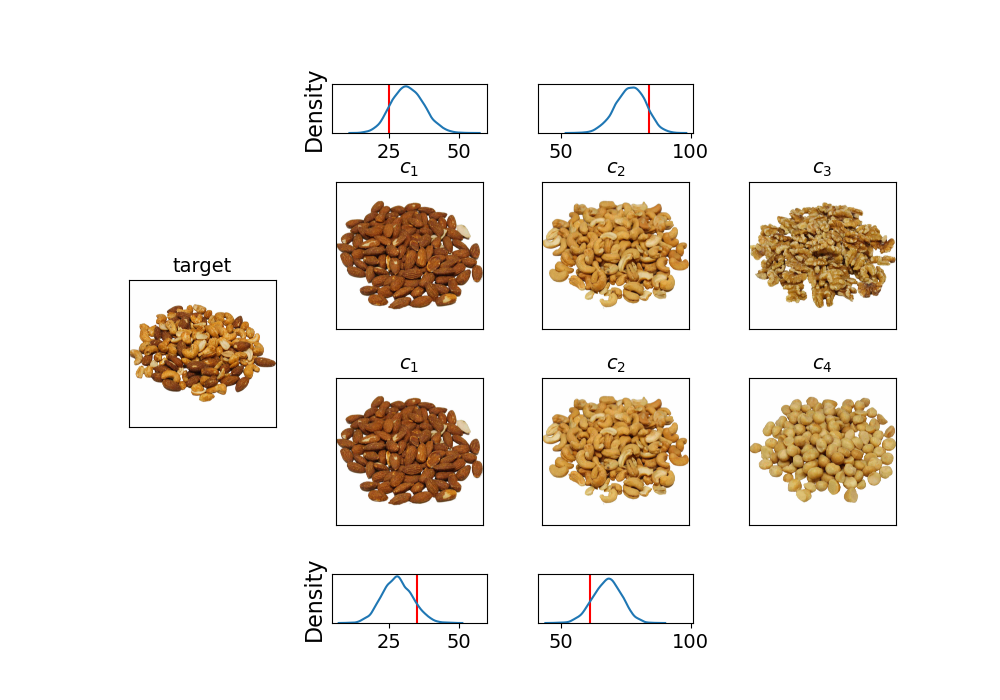}
    \caption{Example of a question pair from the survey. Vertical red line in the plots indicate number of participants selecting that item; blue curve shows the distribution of the (posterior) predicted counts.}
    \label{fig:question_0042}
\end{figure}

\begin{figure}
    \centering
    \includegraphics[width=0.8\linewidth]{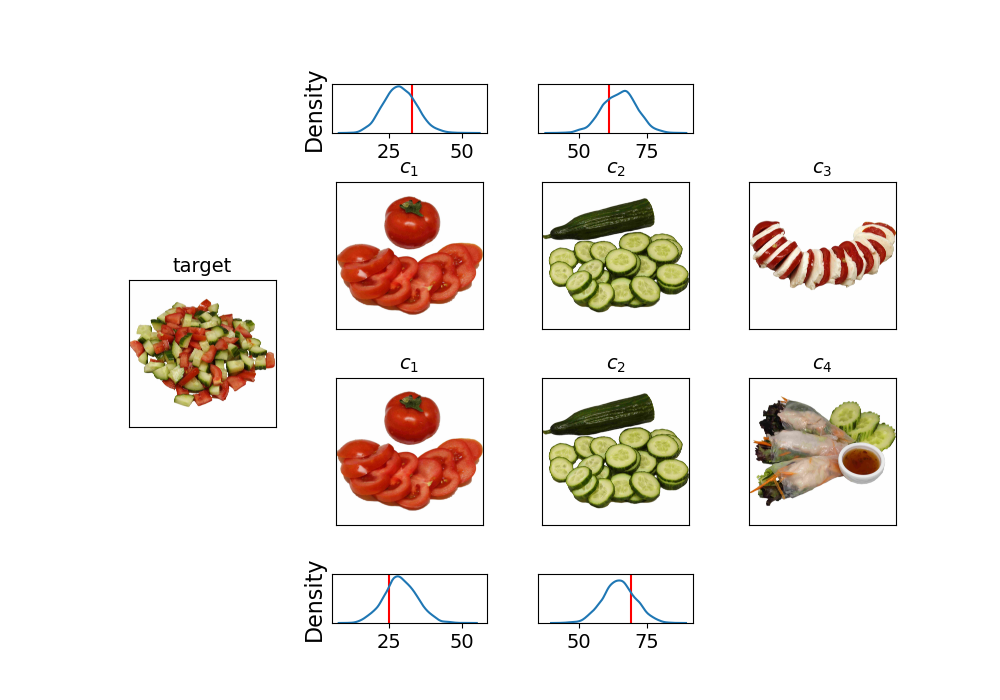}
    \caption{Example of a question pair from the survey. Vertical red line in the plots indicate number of participants selecting that item; blue curve shows the distribution of the (posterior) predicted counts.}
    \label{fig:question_0055}
\end{figure}

\begin{figure}
    \centering
    \includegraphics[width=0.8\linewidth]{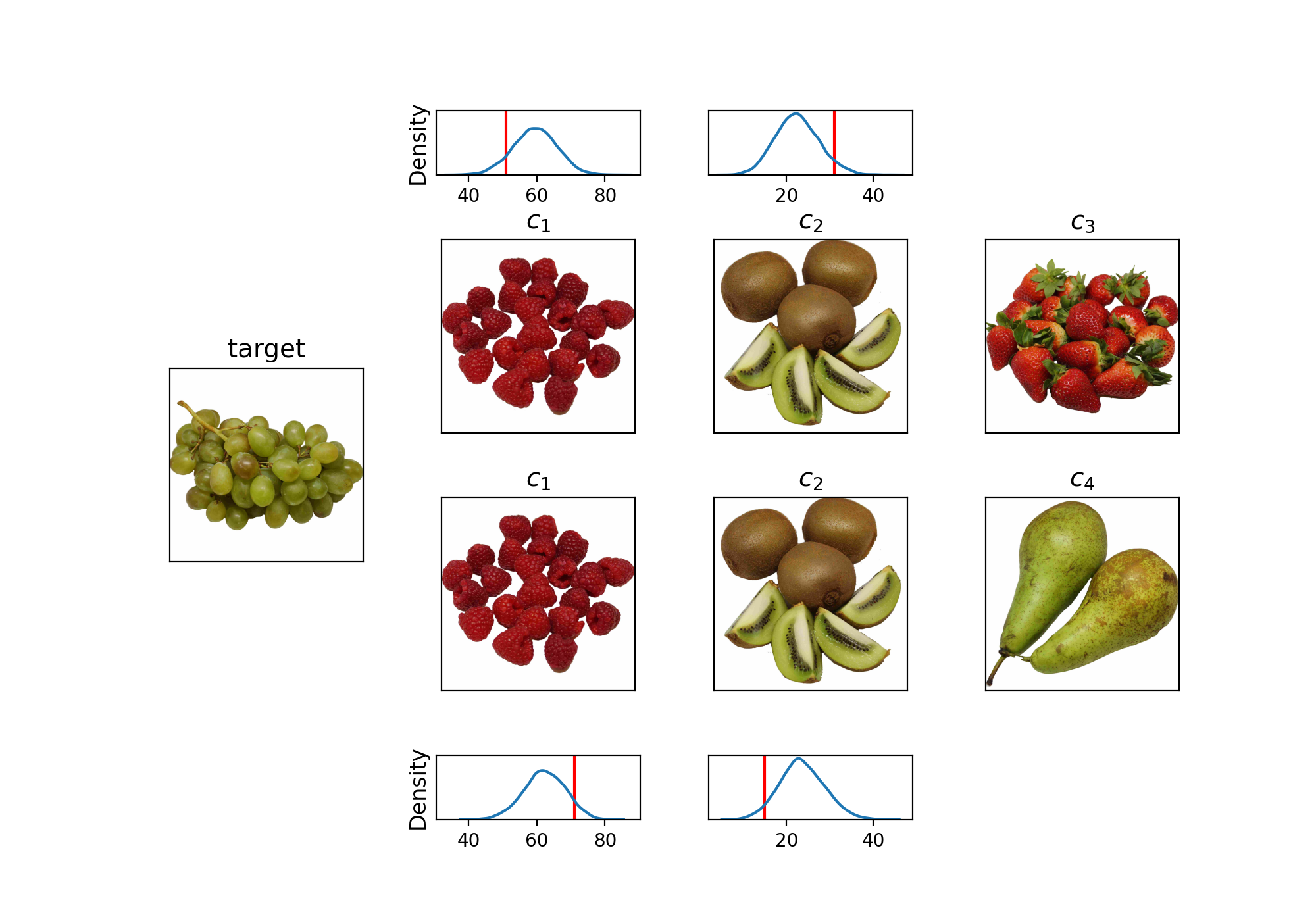}
    \caption{Example of a question pair from the survey. Vertical red line in the plots indicate number of participants selecting that item; blue curve shows the distribution of the (posterior) predicted counts.}
    \label{fig:question_0058dup}
\end{figure}

\begin{figure}
    \centering
    \includegraphics[width=0.8\linewidth]{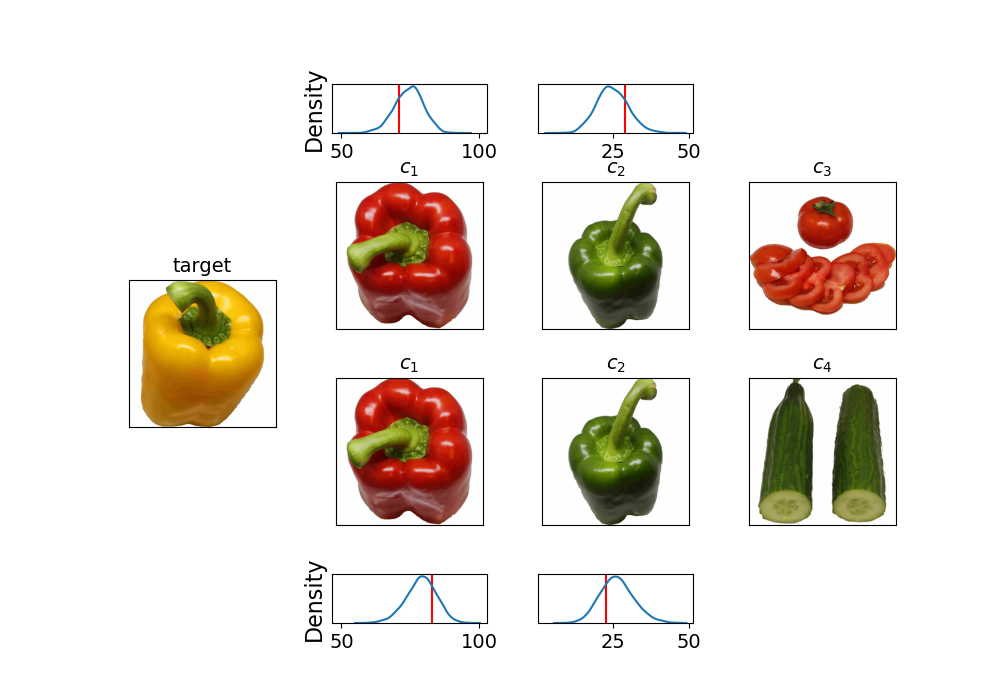}
    \caption{Example of a question pair from the survey. Vertical red line in the plots indicate number of participants selecting that item; blue curve shows the distribution of the (posterior) predicted counts.}
    \label{fig:question_0149}
\end{figure}

\begin{figure}
    \centering
    \includegraphics[width=0.8\linewidth]{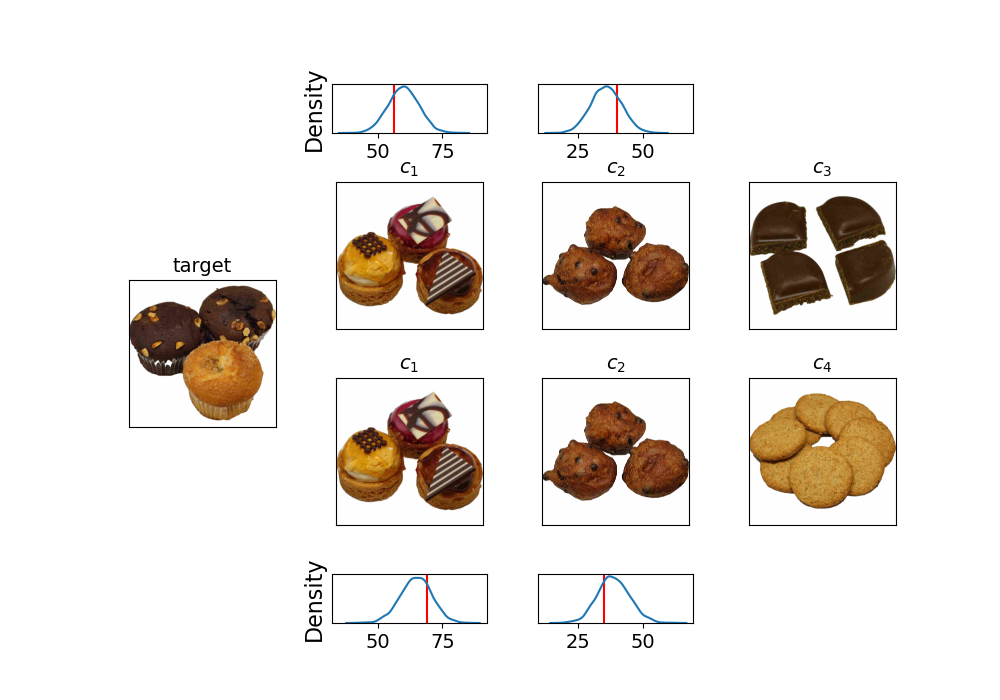}
    \caption{Example of a question pair from the survey. Vertical red line in the plots indicate number of participants selecting that item; blue curve shows the distribution of the (posterior) predicted counts.}
    \label{fig:question_0244}
\end{figure}

\begin{figure}
    \centering
    \includegraphics[width=0.8\linewidth]{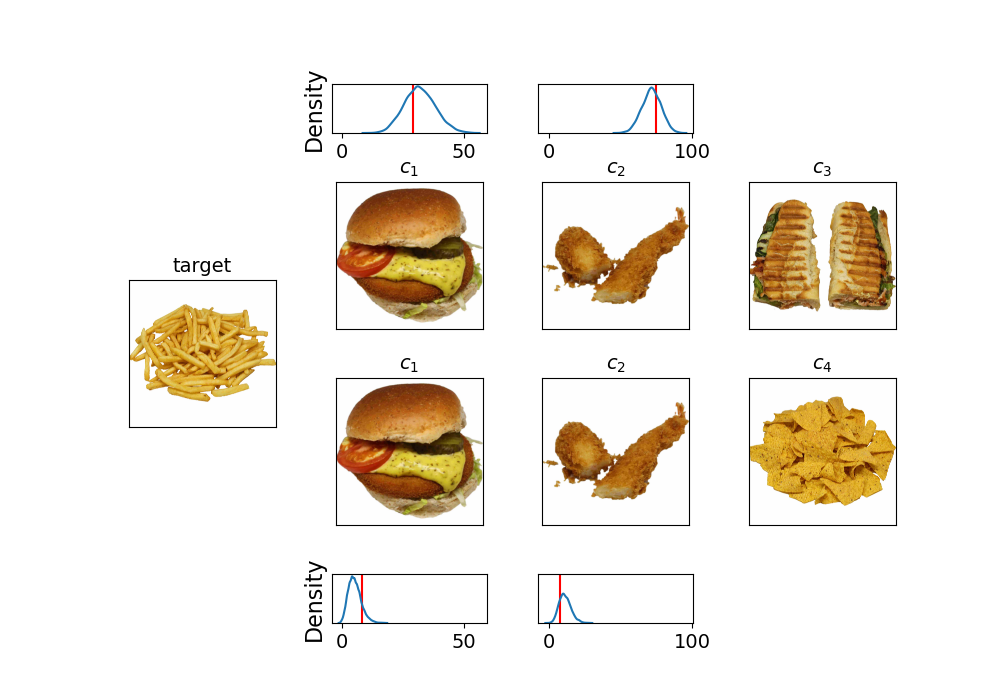}
    \caption{Example of a question pair from the survey. Vertical red line in the plots indicate number of participants selecting that item; blue curve shows the distribution of the (posterior) predicted counts.}
    \label{fig:question_0289}
\end{figure}

\begin{figure}
    \centering
    \includegraphics[width=0.8\linewidth]{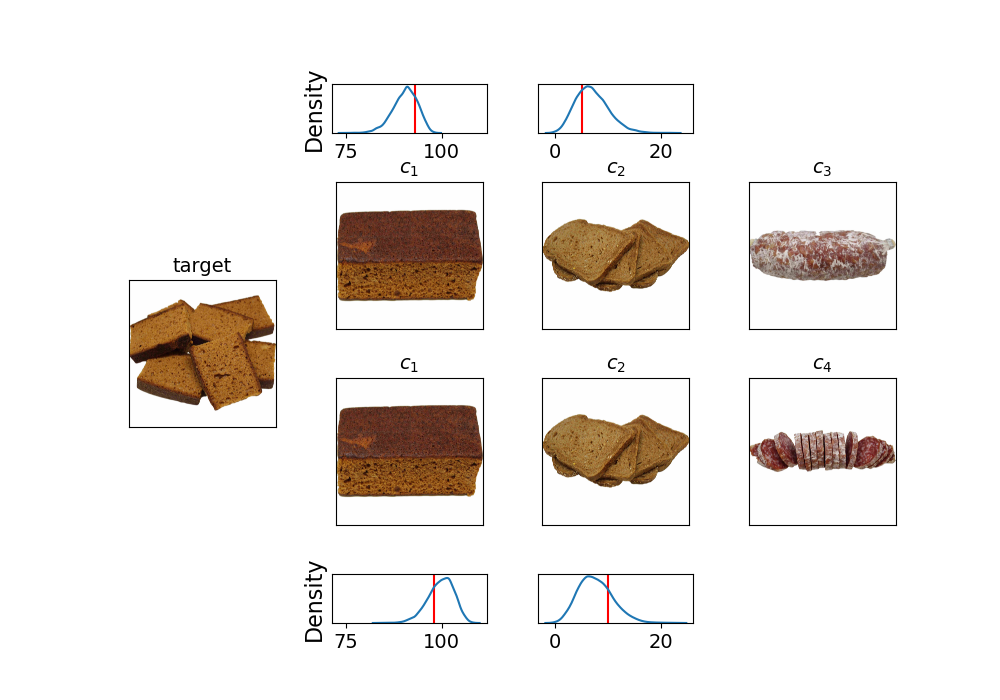}
    \caption{Example of a question pair from the survey. Vertical red line in the plots indicate number of participants selecting that item; blue curve shows the distribution of the (posterior) predicted counts.}
    \label{fig:question_0305}
\end{figure}

\begin{figure}
    \centering
    \includegraphics[width=0.8\linewidth]{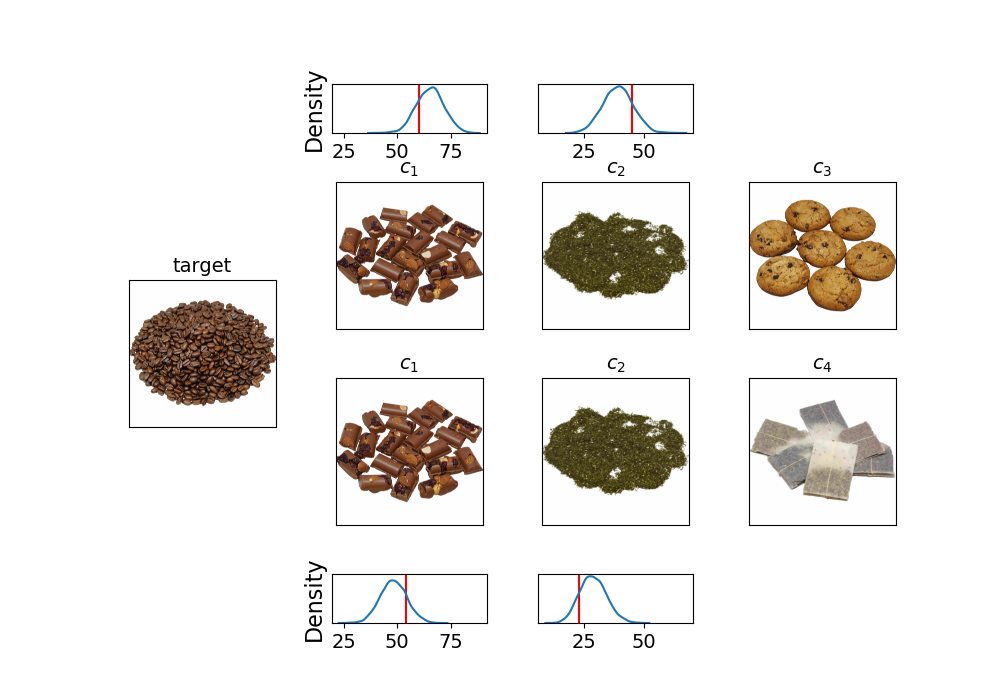}
    \caption{Example of a question pair from the survey. Vertical red line in the plots indicate number of participants selecting that item; blue curve shows the distribution of the (posterior) predicted counts.}
    \label{fig:question_0329}
\end{figure}

\begin{figure}
    \centering
    \includegraphics[width=0.8\linewidth]{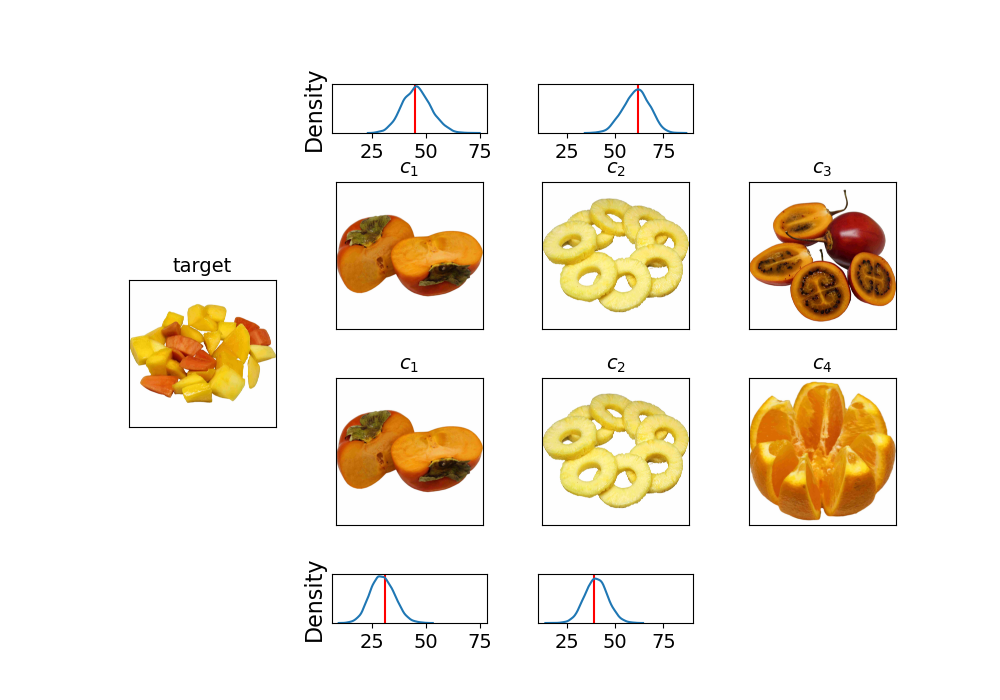}
    \caption{Example of a question pair from the survey. Vertical red line in the plots indicate number of participants selecting that item; blue curve shows the distribution of the (posterior) predicted counts.}
    \label{fig:question_0331}
\end{figure}

\begin{figure}
    \centering
    \includegraphics[width=0.8\linewidth]{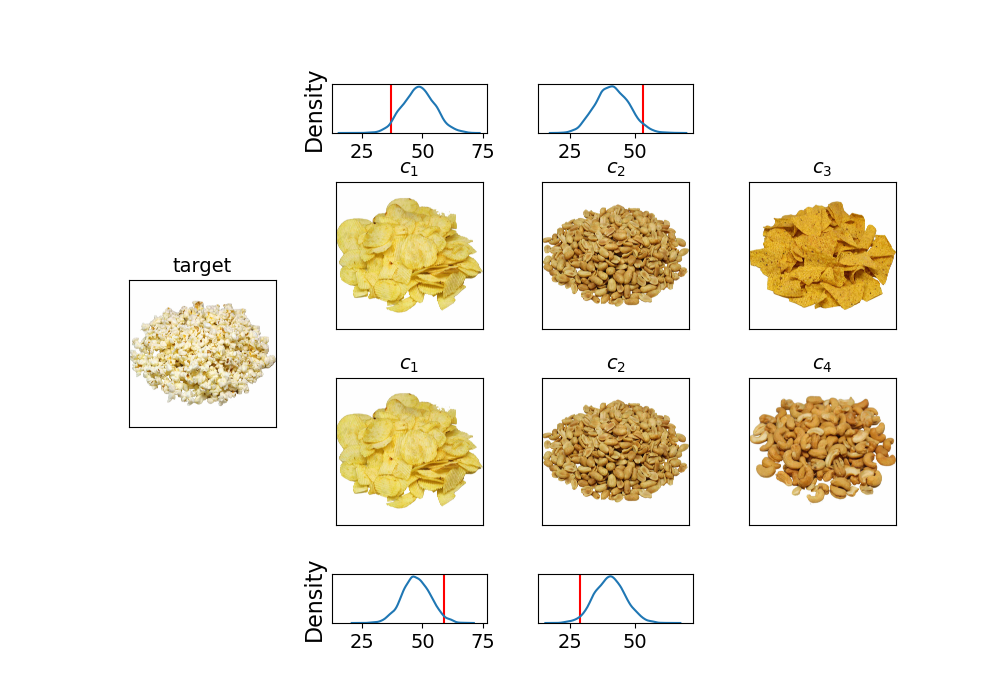}
    \caption{Example of a question pair from the survey. Vertical red line in the plots indicate number of participants selecting that item; blue curve shows the distribution of the (posterior) predicted counts.}
    \label{fig:question_0336}
\end{figure}

\begin{figure}
    \centering
    \includegraphics[width=0.8\linewidth]{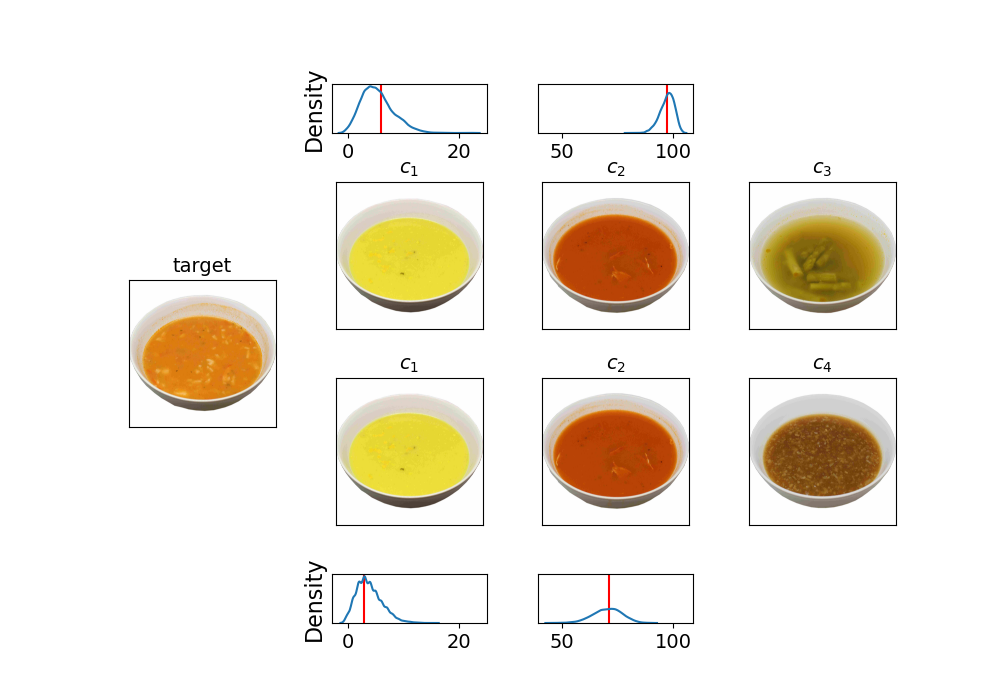}
    \caption{Example of a question pair from the survey. Vertical red line in the plots indicate number of participants selecting that item; blue curve shows the distribution of the (posterior) predicted counts.}
    \label{fig:question_0346}
\end{figure}

\begin{figure}
    \centering
    \includegraphics[width=0.8\linewidth]{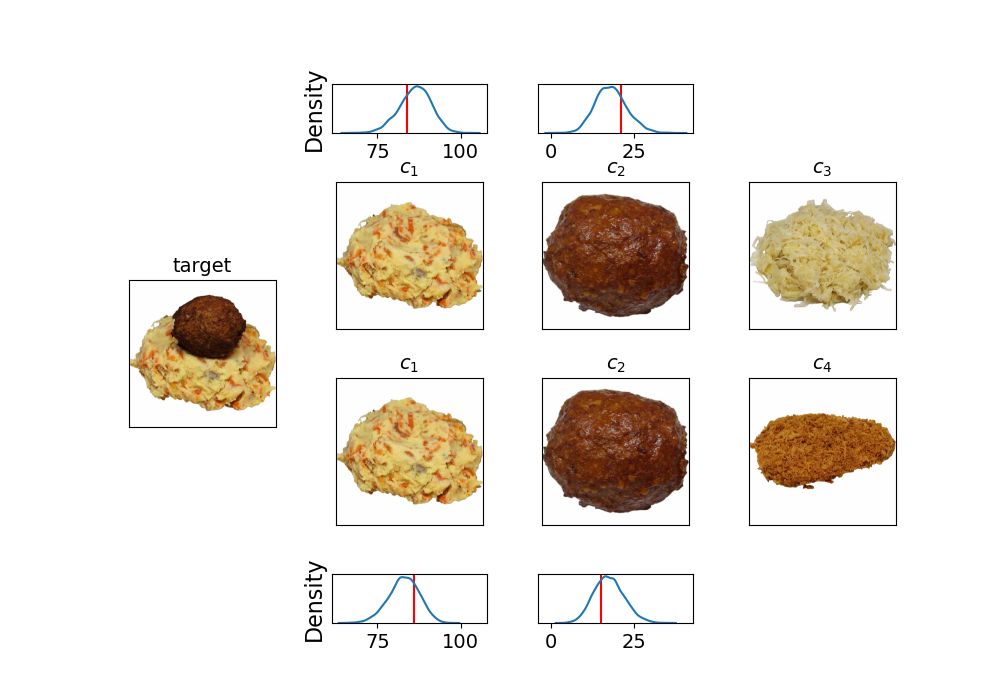}
    \caption{Example of a question pair from the survey. Vertical red line in the plots indicate number of participants selecting that item; blue curve shows the distribution of the (posterior) predicted counts.}
    \label{fig:question_0353}
\end{figure}

\begin{figure}
    \centering
    \includegraphics[width=0.8\linewidth]{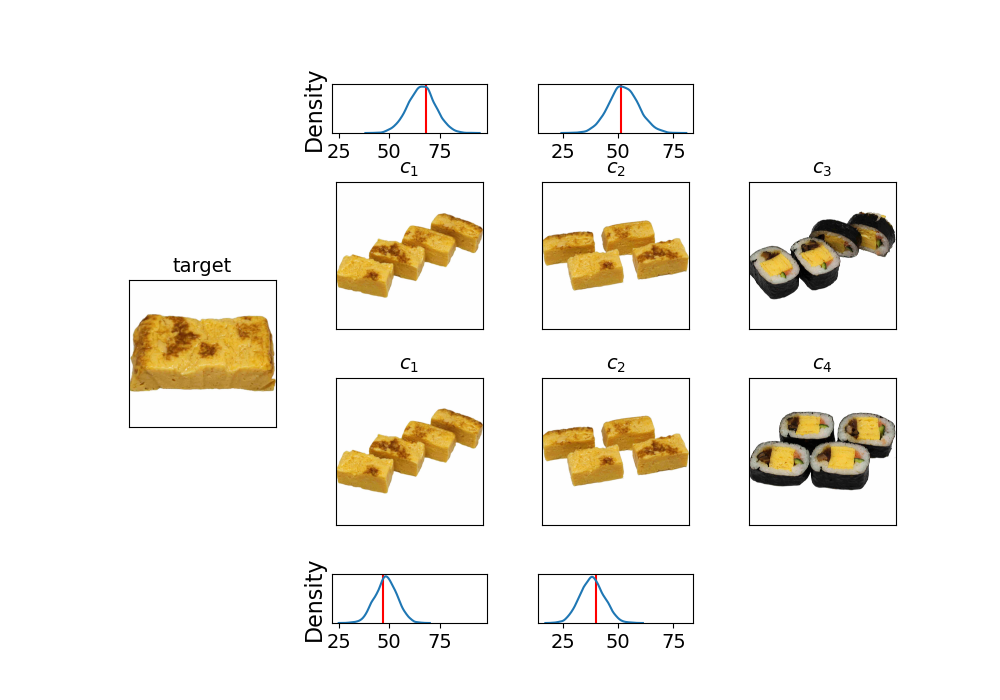}
    \caption{Example of a question pair from the survey. Vertical red line in the plots indicate number of participants selecting that item; blue curve shows the distribution of the (posterior) predicted counts.}
    \label{fig:question_0364}
\end{figure}

\begin{figure}
    \centering
    \includegraphics[width=0.8\linewidth]{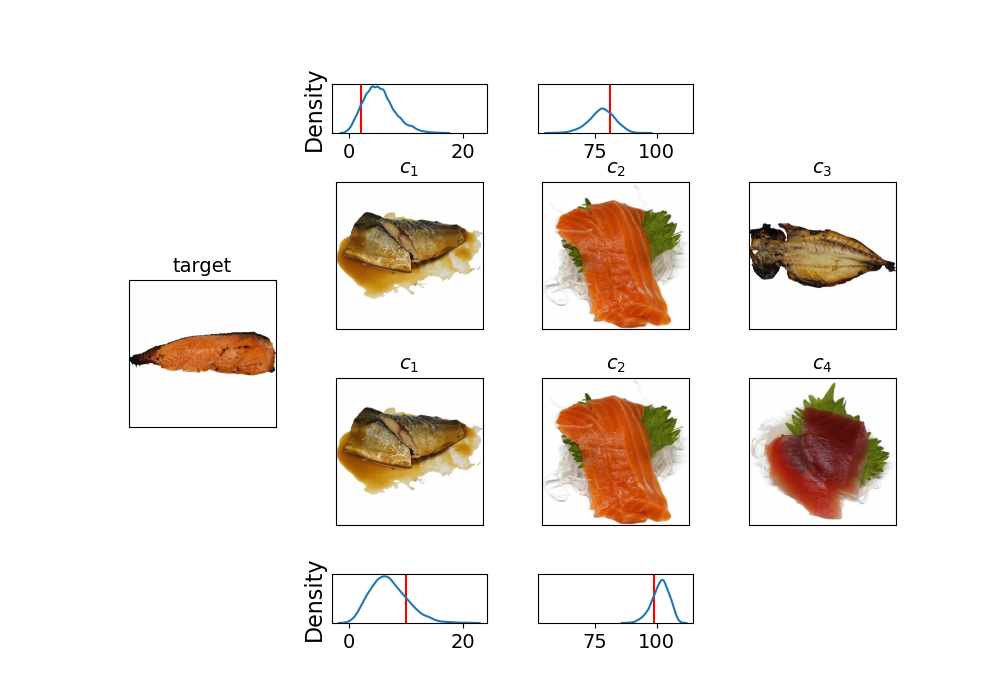}
    \caption{Example of a question pair from the survey. Vertical red line in the plots indicate number of participants selecting that item; blue curve shows the distribution of the (posterior) predicted counts.}
    \label{fig:question_0389}
\end{figure}

\begin{figure}
    \centering
    \includegraphics[width=0.8\linewidth]{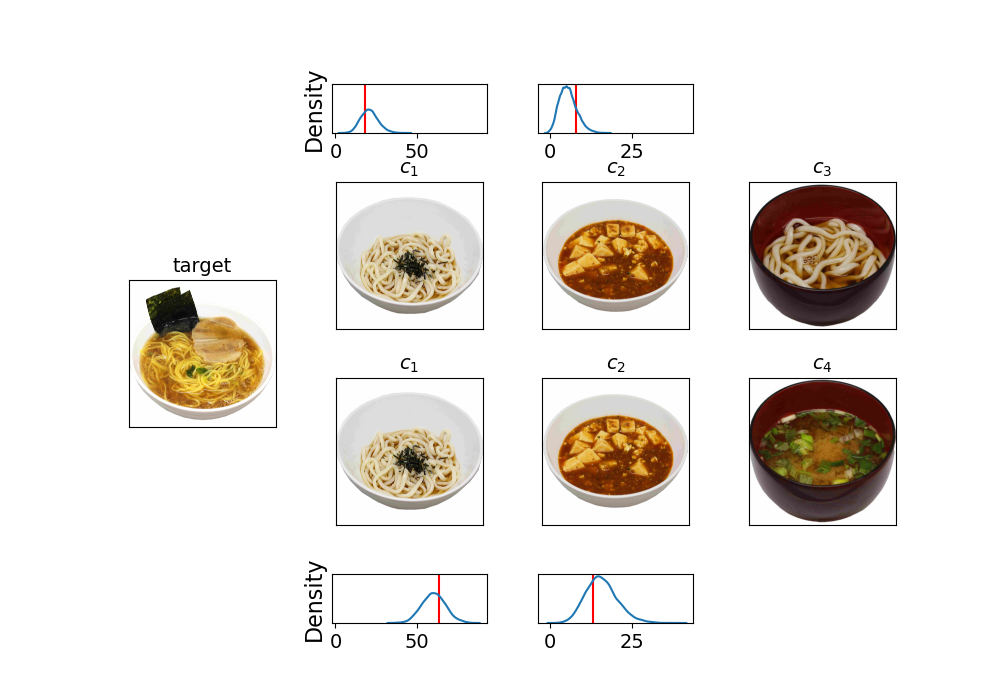}
    \caption{Example of a question pair from the survey. Vertical red line in the plots indicate number of participants selecting that item; blue curve shows the distribution of the (posterior) predicted counts.}
    \label{fig:question_0523}
\end{figure}

\begin{figure}
    \centering
    \includegraphics[width=0.8\linewidth]{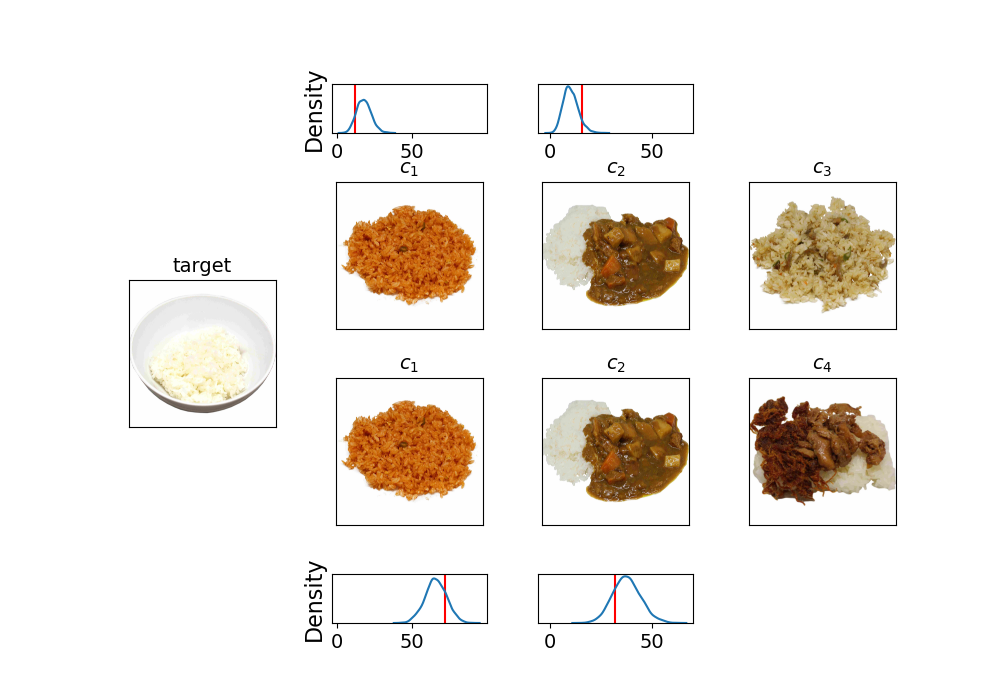}
    \caption{Example of a question pair from the survey. Vertical red line in the plots indicate number of participants selecting that item; blue curve shows the distribution of the (posterior) predicted counts.}
    \label{fig:question_0525}
\end{figure}

\begin{figure}
    \centering
    \includegraphics[width=0.8\linewidth]{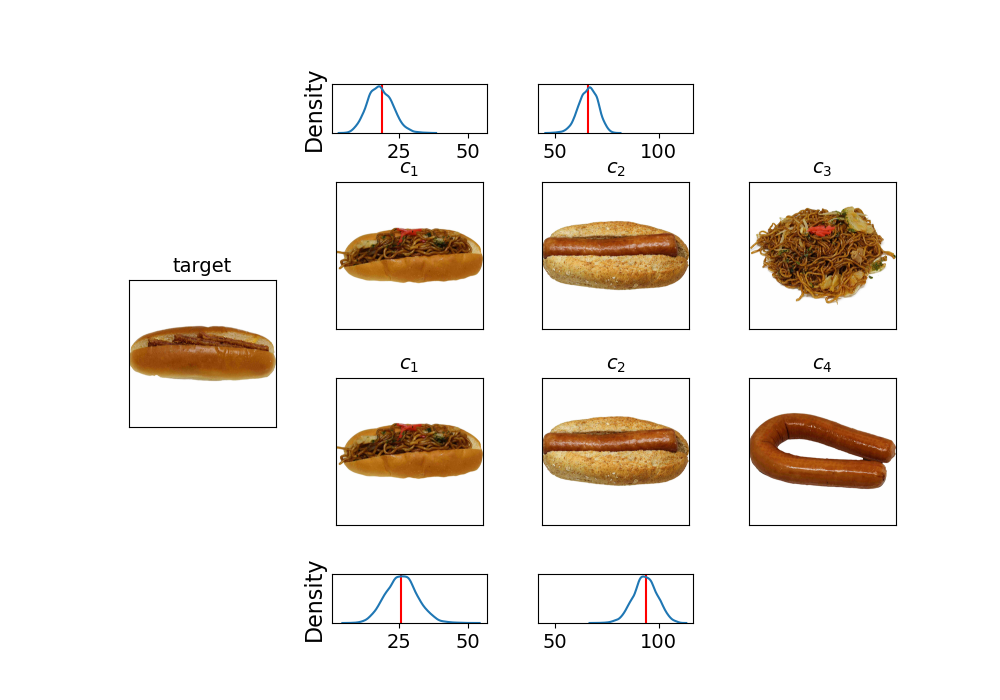}
    \caption{Example of a question pair from the survey. Vertical red line in the plots indicate number of participants selecting that item; blue curve shows the distribution of the (posterior) predicted counts.}
    \label{fig:question_0542}
\end{figure}

\begin{figure}
    \centering
    \includegraphics[width=0.8\linewidth]{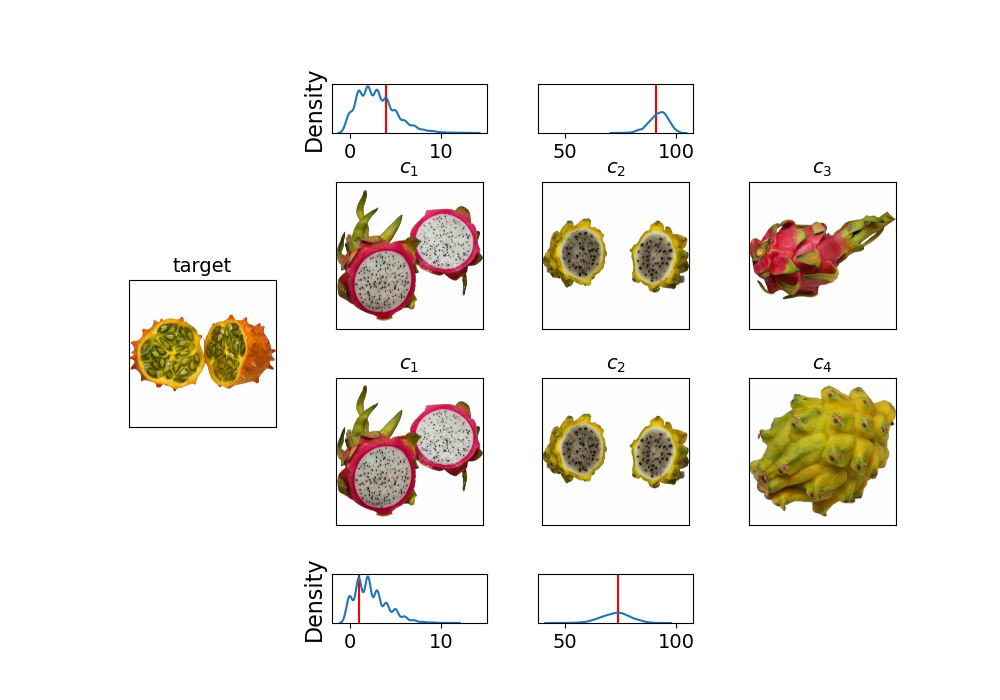}
    \caption{Example of a question pair from the survey. Vertical red line in the plots indicate number of participants selecting that item; blue curve shows the distribution of the (posterior) predicted counts.}
    \label{fig:question_0589}
\end{figure}

\begin{figure}
    \centering
    \includegraphics[width=0.8\linewidth]{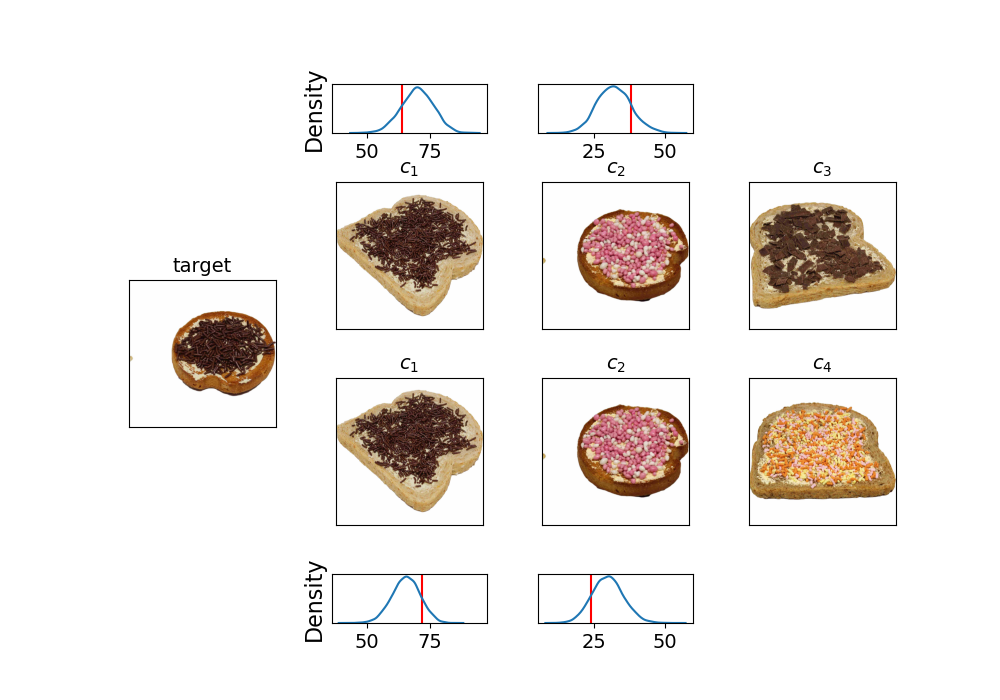}
    \caption{Example of a question pair from the survey. Vertical red line in the plots indicate number of participants selecting that item; blue curve shows the distribution of the (posterior) predicted counts.}
    \label{fig:question_0865}
\end{figure}

\end{document}